\documentclass{article}

\PassOptionsToPackage{numbers,sort&compress}{natbib}
\PassOptionsToPackage{table}{xcolor}

\usepackage[preprint]{neurips_2026}

\usepackage[utf8]{inputenc}
\usepackage[T1]{fontenc}
\usepackage{hyperref}
\usepackage{url}
\usepackage{booktabs}
\usepackage{amsfonts}
\usepackage{amsmath}
\usepackage{amssymb}
\usepackage{nicefrac}
\usepackage{microtype}
\usepackage{xcolor}
\usepackage{graphicx}
\usepackage{multirow}
\usepackage{float}
\usepackage{array}
\usepackage{tabularx}
\usepackage{soul}
\usepackage{algorithm}
\usepackage{algpseudocode}
\newcolumntype{R}[1]{>{\raggedleft\arraybackslash}p{#1}}

\newcolumntype{L}[1]{>{\raggedright\arraybackslash}p{#1}}
\newcolumntype{C}[1]{>{\centering\arraybackslash}p{#1}}
\title{When Molecular Similarity Works: Property Cliffs Reveal Hidden Errors}
\author{
  Di Hu$^{1,\dagger}$,
  Kun Li$^{2,\dagger}$,
  Haojie Rao$^{2}$,
  Longtao Hu$^{2}$,
  Jiameng Chen$^{2}$, and
  Wenbin Hu$^{2,*}$ \\
  $^{1}$School of Economics and Management \\
  $^{2}$School of Computer Science, Wuhan University\\
  Wuhan, China \\
  \texttt{\{dihuleo, likun98, raohaojie, hlt\_2003, jiameng.chen, hwb\}@whu.edu.cn} \\
  $^{\dagger}$These authors contributed equally. \quad
  $^{*}$Corresponding author. \\
  \And
  Yizhen Zheng \\
  Department of Data Science and Artificial Intelligence,\\
  Monash University\\
  Victoria, Australia \\
  \texttt{yizhen.zheng1@monash.edu} \\
  \And
  Jiajun Yu \\
  College of Computer Science and Technology,\\
  Zhejiang University\\
  Hangzhou, China \\
  \texttt{jiajunyu1999@gmail.com} \\
  \And
  Duanhua Cao \\
  School of Life Sciences and Technology,\\
  Tongji University\\
  Shanghai, 200092, China \\
  \texttt{caodh@tongji.edu.cn}
}

\newcommand{\pms}[1]{$\pm${\scriptsize #1}}

\begin{document}
\raggedbottom

\maketitle

\begin{abstract}
Accurate prediction of molecular properties underpins drug discovery and material design, yet even state-of-the-art models remain vulnerable to localized failure modes that aggregate metrics cannot detect. The places where molecular similarity should be most helpful are also places where standard evaluation can be most misleading. Property cliffs expose this gap: structurally similar molecules can still differ sharply in target property, so models with competitive overall performance may fail in high-risk local neighborhoods. To expose and mitigate this failure mode, CliffSplit, a cliff-aware evaluation protocol that constructs locally supported, cliff-exposed test cases, and CliffLoss, a model-agnostic train-only mitigation mechanism for cliff-sensitive errors, are introduced. Experiments on three QM9 targets and three MoleculeNet tasks across five backbones show that CliffSplit reveals at least 15\% higher error in cliff-heavy QM9 regions, while CliffLoss reduces the cliff-to-smooth error gap by up to 30\% on Lipophilicity and improves overall MAE by 9.7\%. Together, these results turn molecular similarity failure from a descriptive anomaly into a benchmarked evaluation problem for molecular machine learning. The code is available at~\url{https://anonymous.4open.science/r/Cliff_Loss}.
\end{abstract}

\section{Introduction}

Drug discovery \citep{icws,zheng2025large,yu2025collaborative} and molecular design \citep{chen2026molevolve,li2025contrastive,li2025drugpilot} address a central human need: finding compounds with desired properties under a massive search space and high experimental cost. Molecular representation learning has become a core capability here~\citep{PCEvo}, with graph neural networks~\citep{yu2025centrality,yu2024kernel}, geometric predictors~\citep{wang2024visnet,liao2023equiformer,schutt2017schnet}, and foundation models~\citep{qin2025moleculeformer,zhou2023unimol} advancing benchmark performance.
Yet model quality is still judged mainly through aggregate metrics, so overall accuracy is often treated as a proxy for reliability. In practice, however, reliability-sensitive tasks such as lead optimization demand that predictions remain trustworthy not only on average but also locally~\citep{li2025bsl}; a molecule predicted accurately in aggregate may still fail precisely where confidence should be highest~\citep{li2026can}.

This tension is sharpened by property cliffs. The similar property principle~\citep{sheridan2004similarity,johnson1990concepts} underlies most molecular predictors: nearby structures should yield nearby values. Property cliffs violate this expectation sharply, as structurally similar molecules can differ dramatically in target property~\citep{maggiora2006outliers,guha2008structure,stumpfe2014activity,stumpfe2022activity}.
The challenge is not merely that some molecules are hard to predict; rather, the hardest failures arise inside apparently well-supported local neighborhoods, where standard evaluation still appears reassuring. A model can achieve low overall error while systematically mispredicting at cliff edges, because aggregate metrics smooth over these localized discontinuities.

\begin{figure}[t!]
\centering
\includegraphics[width=0.98\linewidth]{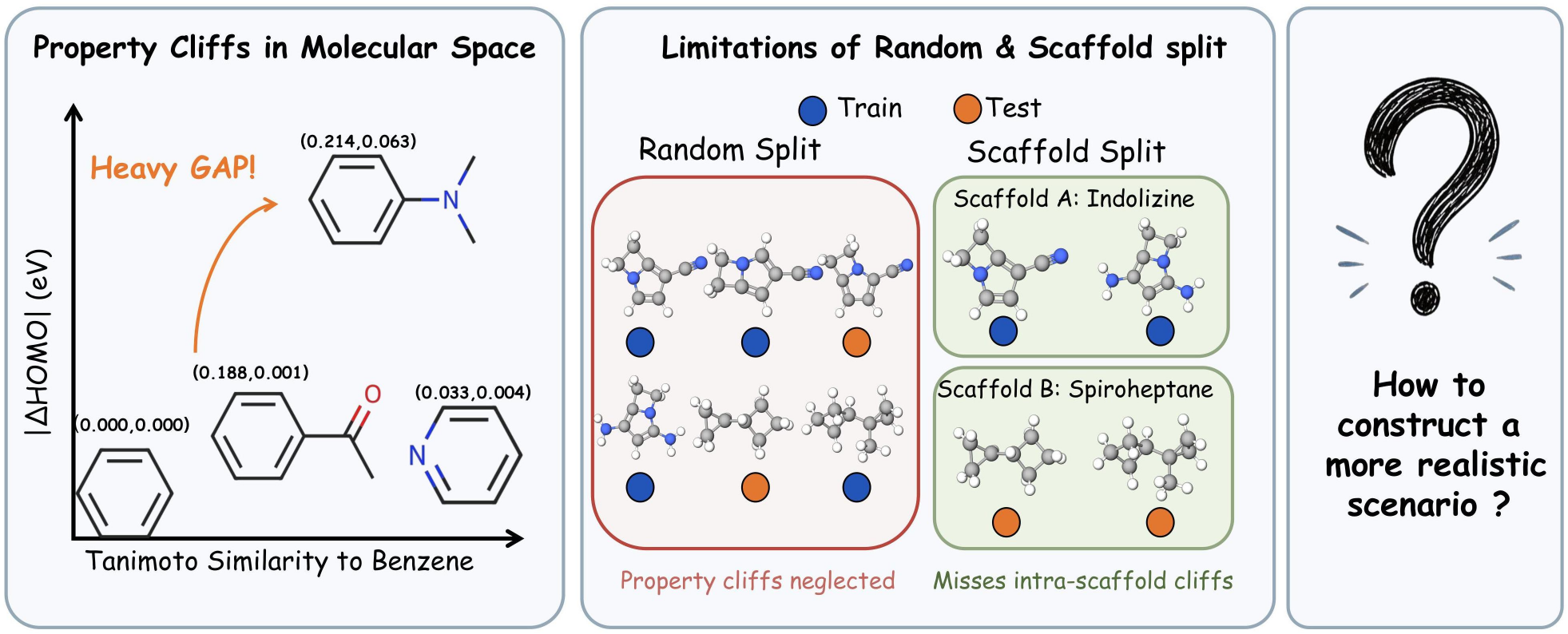}
\caption{Motivation: cliff exposure across split paradigms. Existing splits differ fundamentally in how they treat cross-split cliff exposure, motivating the need for a dedicated cliff-aware evaluation protocol.}
\label{fig:motivation}
\end{figure}

Existing evaluation protocols are not designed to expose this regime. Random split tends to colocate near neighbors across train and test, so cliff-prone regions receive overlapping support and local failure remains hidden: when both ends of a cliff edge are present in training, the model learns a locally smooth fit that appears correct on each molecule individually, and the test set offers no isolated cliff endpoint to expose the discontinuity. Scaffold split tests structural extrapolation at the scaffold level~\citep{bemis1996properties,wu2018moleculenet}, but does not target local property inconsistency under structural support; it separates molecules by chemical skeleton rather than by property gap, so molecules within the same scaffold often share similar property values while true cliff pairs may fall entirely on the same side of the split boundary or into unrelated scaffolds, yielding no explicit cross-split cliff exposure. Figure~\ref{fig:motivation} illustrates this contrast: random split hides cliff errors by placing structurally similar neighbors on both sides of the train/test divide, whereas scaffold split groups molecules by structure rather than by property discontinuity. Neither paradigm constructs cross-boundary cliff exposure as an explicit objective, motivating the need for a dedicated cliff-aware evaluation protocol.

To address this challenge, CliffSplit and CliffLoss, a benchmark-and-mitigation framework for property-cliff reliability in molecular property prediction, are proposed. CliffSplit constructs locally supported yet cliff-exposed train/test partitions to reveal a reliability axis hidden by aggregate metrics, while CliffLoss provides a model-agnostic train-only mitigation mechanism that reweights regression errors by offline cliff severity with adaptive validation feedback. Experiments on three QM9 targets and three MoleculeNet tasks across five backbones show that cliff-heavy QM9 regions incur at least 15\% higher error than smooth regions, and that CliffLoss reduces the cliff-to-smooth error gap by up to 30\% on Lipophilicity while improving overall MAE by 9.7\%. Overall, our contributions can be summarized: (i) a cliff-aware evaluation protocol, CliffSplit, is proposed to construct cliff-exposed train/test partitions and evaluate models via severity-conditioned single-molecule error, exposing a reliability axis invisible to aggregate metrics; (ii) a model-agnostic loss plugin, CliffLoss, is introduced to reweight regression errors by offline cliff severity with adaptive control, reducing cliff-region degradation without architectural modification; (iii) Experiments on QM9 and MoleculeNet across five backbones show that CliffSplit exposes systematic severity-dependent degradation, with at least 15\% higher error in cliff-heavy QM9 regions, while CliffLoss reduces the Lipophilicity gap by up to 30\% without sacrificing overall accuracy.

\section{Related Work}

The similar property principle~\citep{sheridan2004similarity} has long guided chemoinformatics and continues to shape modern molecular predictors, which inherit an implicit smoothness bias~\citep{johnson1990concepts,gilmer2017neural}. Activity cliffs, sharp local discontinuities where structurally similar molecules exhibit large property differences, have been extensively documented in bioactivity landscapes, and ML models have been shown to be least accurate precisely in these regions~\citep{maggiora2006outliers,guha2008structure,stumpfe2014activity,stumpfe2022activity,dimova2016advances,sheridan2013random}. That literature established the chemical reality of cliff-prone error, but its main focus remains cliff detection and SAR interpretation rather than benchmark design for general molecular property prediction.

This shift makes evaluation protocols central. Modern molecular property prediction spans diverse model families~\citep{yang2019analyzing,schutt2017schnet,wang2024visnet,aykent2025gotennet,ross2022molformer,qin2025moleculeformer,zhou2023unimol} and standard benchmarks such as QM9 and MoleculeNet have standardized evaluation across quantum-chemical and experimental properties~\citep{ramakrishnan2014quantum,wu2018moleculenet,bemis1996properties}. Yet dominant practice still relies on MAE- or RMSE-style pointwise summaries under random or scaffold splits, which do not directly test whether error concentrates in high-similarity neighborhoods where the similarity principle should be most informative. The missing component is therefore not another model family, but a released protocol that exposes cliff-heavy local neighborhoods across the train/test boundary.

Once the problem is framed as reliability rather than average accuracy, it connects to robustness and hard-case learning, including example reweighting, long-tail aware methods, hard-region reweighting, distributionally robust optimization, invariant learning, predictive uncertainty, out-of-distribution (OOD) and graph OOD benchmarks, calibration, and relation-aware objectives~\citep{ren2018learning,menon2021longtail,lin2017focal,sagawa2020distributionally,arjovsky2020invariant,gal2016dropout,lakshminarayanan2017simple,koh2021wilds,guo2017calibration,chen2022good,hu2020ogb,liu2023topology,li2025contrastive}. These directions are relevant, but they largely assume that difficult regions are already identified. What remains underdeveloped is the benchmark layer itself: a protocol that constructs cliff-exposed test regions, induces molecule-level severity from local cliff structure, and evaluates standard single-molecule predictors by severity-conditioned error rather than aggregate accuracy alone.

\section{Cliff Benchmark}
\label{sec:cliff_benchmark}

Standard molecular benchmarks report aggregate pointwise metrics such as MAE, RMSE, or $R^2$ under random or scaffold split, but they do not test whether error concentrates in the high-similarity neighborhoods where molecular similarity should help most. CliffSplit addresses this gap by exposing locally inconsistent neighborhoods across the train/test boundary and evaluating severity-conditioned single-molecule MAE on cliff-exposed test molecules.

\subsection{Problem Setup}

Let the molecular dataset be
\begin{equation}
    \mathcal{D} = \{(x_i, y_i)\}_{i=1}^{N},
    \label{eq:dataset}
\end{equation}
and let a single-molecule predictor be $f: \mathcal{X} \to \mathbb{R}$ with predictions $\hat{y}_i = f(x_i)$.
\begin{equation}
    f: \mathcal{X} \rightarrow \mathbb{R}, \qquad \hat{y}_i = f(x_i),
    \label{eq:predictor}
\end{equation}

For any pair $(x_i, x_j)$ define structural similarity $s_{ij}$ and absolute property difference
\begin{equation}
    \Delta y_{ij}=|y_i-y_j|.
    \label{eq:delta_y}
\end{equation}
These pair-level quantities enter the benchmark in two places: they determine which edges form the cliff graph, and once the split is fixed they induce molecule-level severity through training-side neighborhoods. To identify difficult pairs, a high-risk pair score is defined that jointly considers structural similarity and property divergence via the scale-free joint score
\begin{equation}
    C_{ij} = s_{ij}^{\alpha} \cdot r_{ij}^{\beta},
    \label{eq:cliffscore}
\end{equation}
where $r_{ij}\in[0,1]$ is the global rank percentile of $\Delta y_{ij}$ among all candidate pairs; large $C_{ij}$ identifies pairs that are both highly similar and far apart in property space. Thresholding at a top-$k$ quantile $\tau_C$ defines the raw cliff-candidate set
\begin{equation}
    E_{\mathrm{cliff}}^{\mathrm{raw}}=\{(i,j)\in\mathcal{P}\mid C_{ij}\ge\tau_C\}.
    \label{eq:raw_cliff}
\end{equation}
This candidate set is then refined into the final cliff graph as described in Section~\ref{subsec:benchmark}.

\subsection{Benchmark Construction}
\label{subsec:benchmark}

CliffSplit constructs a cliff graph from high-scoring pairs (Section~\ref{eq:cliffscore}), partitions it under a hard constraint, and assigns training-induced severity groups for evaluation. All construction uses ground-truth labels offline; no model inference is required.

Pairs exceeding threshold $\tau_C$ on the CliffScore form $E_{\mathrm{cliff}}^{\mathrm{raw}}$, as given in Equation~\ref{eq:raw_cliff}. A degree-capped greedy selection removes hub-dominated edges, yielding
\begin{equation}
    G_{\mathrm{cliff}} = (V, E_{\mathrm{cliff}}),
    \label{eq:cliff_graph}
\end{equation}
where $V$ is the set of molecules with at least one retained cliff edge. The graph-based split is computed component-wise under a hard constraint forbidding any cliff edge from having both endpoints assigned to test, while maximizing cross-split cliff exposure, defined as the fraction of cliff edges straddling the train/test boundary. Bipartite components admit exact two-color assignment; non-bipartite components require constrained optimization. Partition quality is measured by induced coverage
\begin{equation}
    \mathrm{Cov}(V_{\mathrm{tr}}, V_{\mathrm{te}})
    =
    \frac{\left|\left\{(i,j)\in E_{\mathrm{cliff}} \mid (i\in V_{\mathrm{tr}},\ j\in V_{\mathrm{te}})\ \mathrm{or}\ (i\in V_{\mathrm{te}},\ j\in V_{\mathrm{tr}})\right\}\right|}{|E_{\mathrm{cliff}}|}.
    \label{eq:coverage}
\end{equation}

Validation is sampled uniformly from the training pool after fixing train/test. Each edge $(i,j)\in E_{\mathrm{cliff}}$ carries a quartile label
\begin{equation}
    q_{ij} = \left\lfloor 4\cdot\frac{\mathrm{rank}((i,j))}{|E_{\mathrm{cliff}}|}\right\rfloor + 1,
    \label{eq:pair_quartile}
\end{equation}
based on descending CliffScore rank, where Q1 denotes the lowest-severity quartile and Q4 the highest. At molecule level, each test molecule's severity is induced exclusively from training-side neighbors:
\begin{equation}
    Q(v_\alpha) = \max_{(v_\alpha, v_k)\in E_{\mathrm{cliff}},\ v_k\in V_{\mathrm{tr}}} q_{\alpha k},
    \label{eq:mol_severity}
\end{equation}
so no information derived from the test set itself enters the assignment. The resulting molecule-level groups Q1--Q4 are therefore not arbitrary bins over individual molecules but severity strata induced from local cliff structure under train-side support. The complete construction procedure is summarized in Algorithm~\ref{alg:cliffsplit} in Appendix~\ref{sec:app_method_details}.

\subsection{Why CliffSplit Is Hard}

Using QM9's three quantum-chemical properties (GAP, HOMO, LUMO) as running example, what makes CliffSplit distinct from conventional benchmarks is characterized by three facts. Three facts jointly establish that it measures a targeted failure mode rather than generic OOD error.

\begin{figure}[t!]
\centering
\includegraphics[width=0.86\linewidth]{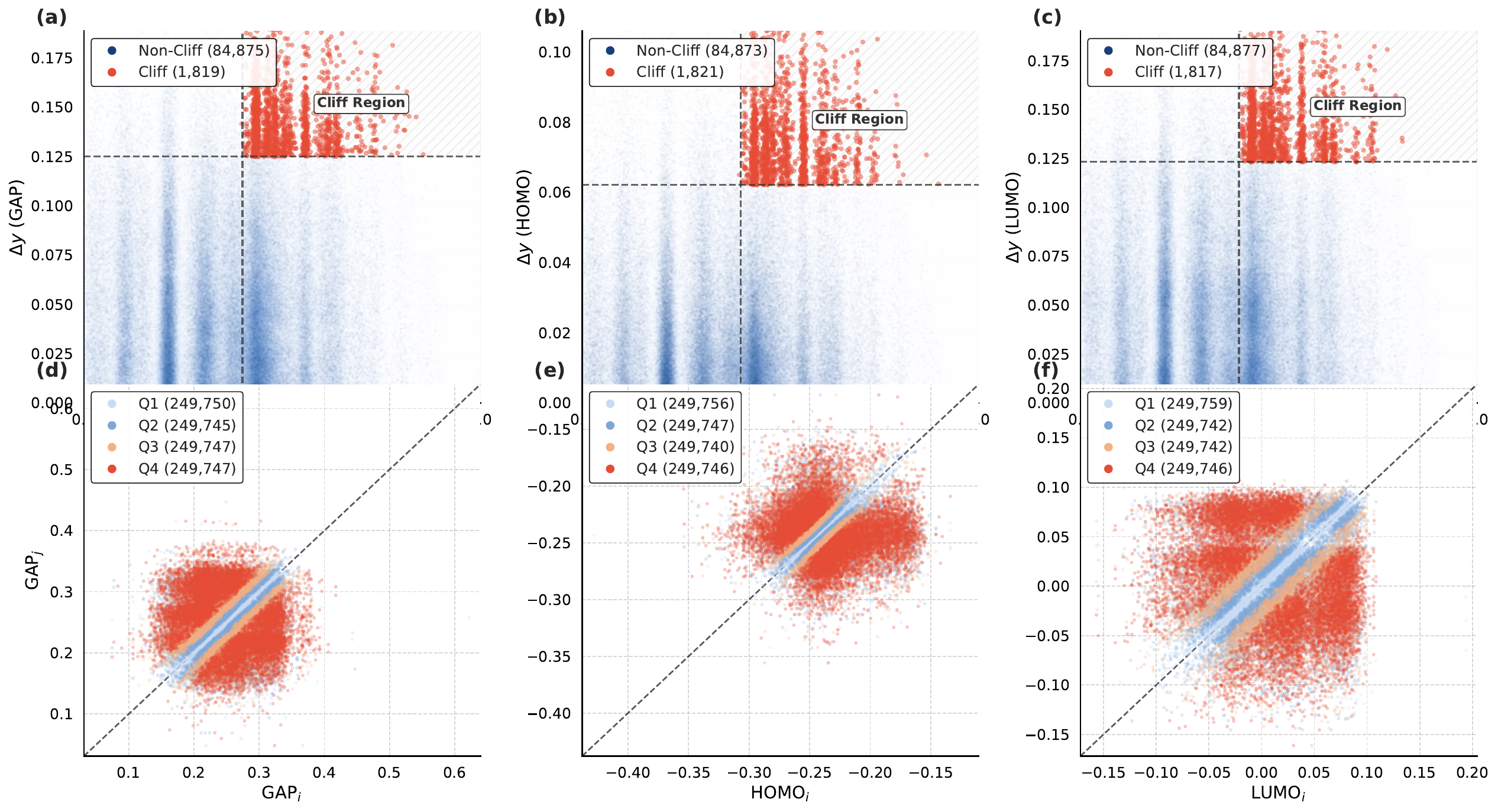}
\caption{Property-cliff regions and severity geometry. (a)(b)(c) Zoomed high-similarity regime for GAP, HOMO, and LUMO: blue density denotes the smooth-response majority; red points mark cliff pairs in the high-similarity, high-divergence corner. (d)(e)(f) Pair endpoint distribution under the Q1--Q4 labels: higher quartiles move progressively away from the diagonal $y_i = y_j$, giving a direct geometric interpretation of cliff severity in property space.}
\label{fig:property_cliff_region}
\end{figure}

The targeted cliff regime arises when structurally similar molecules exhibit large property differences, a failure mode hidden under aggregate metrics, as shown in Figure~\ref{fig:property_cliff_region}. For all three properties, the dominant molecular mass occupies a smooth-response region shown in blue, while a structured minority occupies the high-similarity, high-$\Delta y$ corner shown in red. Standard evaluation averages over both regions, so error in the cliff corner is diluted by the smooth majority. CliffSplit targets this corner explicitly.

The degradation under SCOPE-Bench is not simply caused by marginal extrapolation. Train and test distributions remain broadly overlapping, with normalized Wasserstein-1 distances $W_1/\sigma$ between 0.47 and 0.56 and KDE overlaps above 0.57 across all tasks (\mbox{Appendix~\ref{fig:split_kde}}). Every test molecule retains dense nearest-training neighbor support: median Tanimoto stays from 0.571 to 0.591, with ${>}80\%$ above 0.5 (\mbox{Appendix~\ref{fig:nearest_train_similarity}}). The benchmark, therefore, tests reliability where structural neighbors exist but become locally inconsistent, not in regions devoid of training coverage.

CliffSplit actively reallocates cliff edges toward the train/test boundary: among all cliff edges incident on a test molecule, 88.6\% to 89.0\% connect to the training set and only 11.0\% to 11.4\% to validation. On average, each test molecule retains from 10.6 to 16.9 training-side cliff partners, as shown in Table~\ref{tab:cliffsplit_qm9}. This cross-split placement surfaces cliff phenomena for evaluation under dense local support rather than hiding them inside a single partition.

\subsection{CliffSplit Benchmark Analysis}

Having established that CliffSplit is hard for a principled reason, the next question is whether the resulting difficulty is structured: can the benchmark decompose test error along a stable severity axis rather than reporting a single aggregate number? The pair-level quartile labels $q_{ij}\in\{1,2,3,4\}$ are defined in Equation~\ref{eq:pair_quartile} and partition cliff edges by descending CliffScore rank. Each test molecule inherits its severity group from Equation~\ref{eq:mol_severity}, yielding molecule-level partitions
\begin{equation}
    V_q = \bigl\{v_\alpha \in V_{\mathrm{te}} \;\big|\; Q(v_\alpha) = q\bigr\}, \quad q \in \{1,2,3,4\},
    \label{eq:quartile_groups}
\end{equation}
where $|V_1| + |V_2| + |V_3| + |V_4| = |V_{\mathrm{te}}|$. Per-quartile mean absolute error and the severity ratio are then defined as
\begin{align}
    \bar{e}_q &= \frac{1}{|V_q|} \sum_{i \in V_q} \bigl|\hat{y}_i - y_i\bigr|,
    \label{eq:mae_quartile}\\[4pt]
    R &= \frac{\bar{e}_4}{\bar{e}_1}.
    \label{eq:severity_ratio}
\end{align}
Values of $R > 1$ indicate that the highest-severity quartile incurs proportionally larger error than the easiest quartile, quantifying cliff-induced error amplification. Mean $\Delta y$ increases monotonically from Q1 to Q4 with Q4/Q1 ratios of 2.52 (GAP), 2.73 (HOMO), and 2.70 (LUMO); mean Tanimoto rises from ${\sim}0.04$ to ${\sim}0.11$, so higher quartiles are simultaneously more similar and more property-separated, as shown in Table~\ref{tab:cliffsplit_qm9}. In property space, higher quartiles move progressively away from the diagonal $y_i = y_j$, as shown in the bottom row of Figure~\ref{fig:property_cliff_region}, confirming that severity corresponds to stronger local contradiction between structural similarity and property value. Thus, CliffSplit evaluates not only overall error but also how models degrade under increasing cliff severity, a diagnostic dimension invisible to aggregate metrics.

\section{Method}

The benchmark makes property-cliff failure observable, but evaluation alone does not reduce it. CliffLoss is introduced as a train-only, model-agnostic loss plugin that reweights regression errors by precomputed cliff severity while leaving the backbone unchanged. CliffLoss shares the same offline cliff graph with CliffSplit but applies it only during training, so no test targets, learned cliff detector, or inference-time operations are involved.

\begin{figure}[ht!]
\centering
\includegraphics[width=0.75\linewidth]{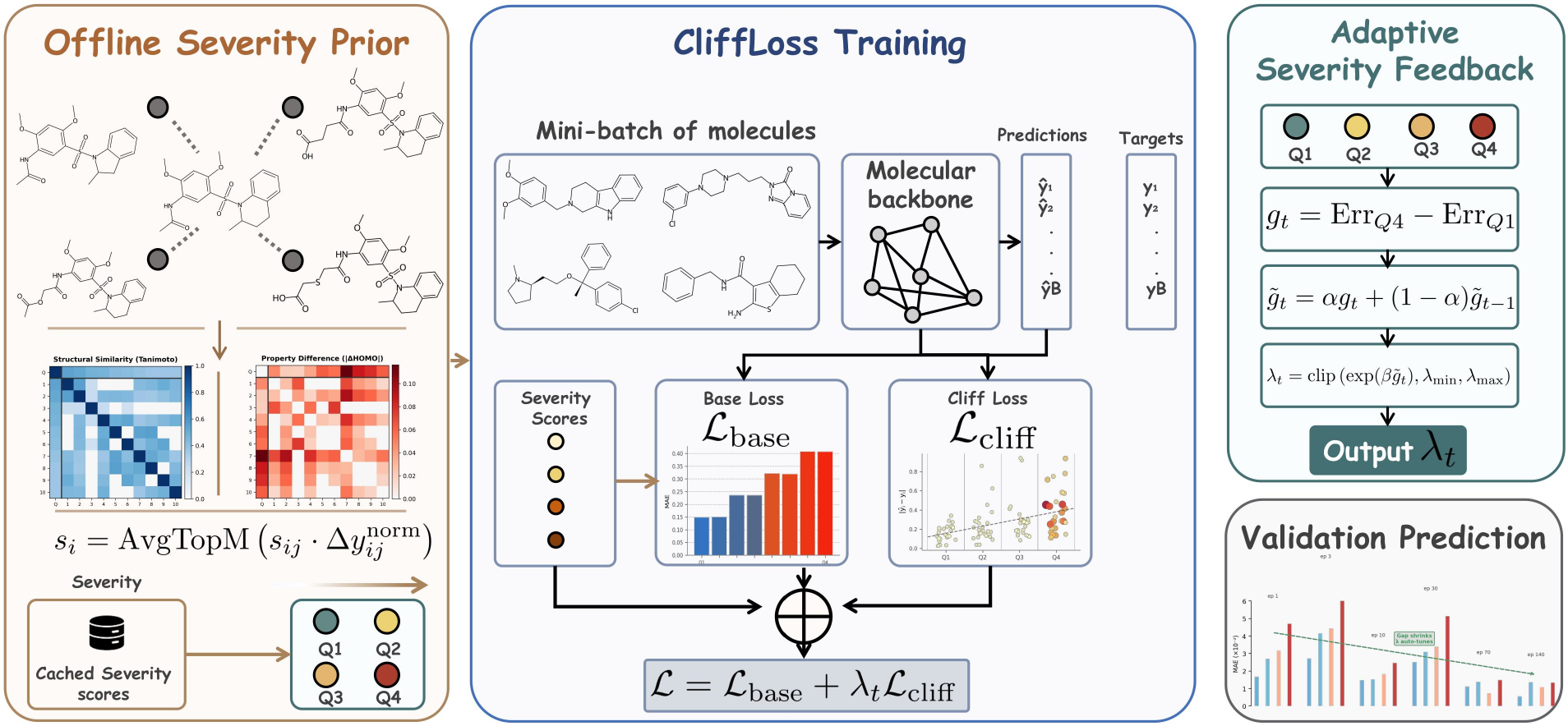}
\caption{CliffLoss training pipeline. Offline cliff scores are precomputed and fixed. During training, the base loss and the score-weighted cliff loss are combined with an adaptive weight \mbox{\(\lambda_t\)} derived from validation severity diagnostics. At inference, only the backbone is used.}
\label{fig:cliffloss-mechanism}
\end{figure}

\subsection{Overview}

Cliff-heavy molecules form a training minority but dominate reliability failure; uniform loss dilutes their gradient signal. CliffLoss corrects this at the loss level without modifying the backbone architecture. For molecule \(i\) with offline severity score \(s_i\), the total objective is
\begin{equation}
    \mathcal{L} = \underbrace{\frac{1}{N}\sum_{i=1}^{N} |\hat{y}_i - y_i|}_{\mathcal{L}_{\mathrm{base}}} + \lambda_t \cdot \underbrace{\frac{1}{N}\sum_{i=1}^{N} s_i \cdot |\hat{y}_i - y_i|}_{\mathcal{L}_{\mathrm{cliff}}},
    \label{eq:l_total}
\end{equation}
where \(\lambda_t\) is adaptively updated from validation severity diagnostics and the backbone \(f_\theta\) is unchanged. The complete training procedure is described in Algorithm~\ref{alg:cliffloss} in Appendix~\ref{sec:app_method_details}.

\subsection{Offline Cliff Scores}

CliffLoss assigns a per-molecule severity score to every training sample before training begins, constructed offline from raw pair-level statistics via localized neighbor aggregation.

Neighbor collection. For each molecule \(i\) in any of the training, validation, or test splits, its \(\tau\)-similar neighbors are collected from the training set as reference, retaining at most \(K\) pairs with the highest Tanimoto similarity:
\begin{equation}
    \mathcal{N}_K(i) = \{(j, s_{ij}, \Delta y_{ij}) \mid j \in V_{\mathrm{tr}},\ s_{ij} \ge \tau,\ i \neq j,\ |\mathcal{N}_K(i)| \le K\},
    \label{eq:neighbors}
\end{equation}
where $V_{\mathrm{tr}}$ is the training set, $s_{ij}$ is Tanimoto similarity, and $\Delta y_{ij}=|y_i-y_j|$. For val/test molecules the reference pool remains $V_{\mathrm{tr}}$, so severity is always grounded in training-side neighborhoods.

Severity score definition. From the collected neighbors, each molecule receives a scalar severity score, defined as
\begin{equation}
    s_i = \frac{1}{M}\sum_{(j,\cdot,\cdot)\in \mathcal{N}_M(i)} s_{ij} \cdot \frac{\Delta y_{ij}}{q_{0.95}},
    \label{eq:score_def}
\end{equation}
where \(\mathcal{N}_M(i) \subseteq \mathcal{N}_K(i)\) retains only the \(M\) pairs with the largest normalized strength \(s_{ij} \cdot \Delta y_{ij}\), and \(q_{0.95}\) is the 95th-percentile of all collected \(\Delta y_{ij}\) values over the training set, serving as a property-specific normalization constant. A large \(s_i\) indicates that molecule \(i\) lies in a locally supported yet property-inconsistent region, so its prediction error should carry larger optimization weight. The scores are computed offline and reused across all backbones, keeping the training signal backbone-independent and directly comparable across models.

\subsection{Adaptive Cliff Weight}

A fixed cliff weight faces a basic tradeoff: if it is too small, the correction is negligible; if it is too large, the model can over-focus on cliff-heavy neighborhoods and degrade smooth-region accuracy. CliffLoss resolves this by adapting \(\lambda_t\) online using the validation-time error gap between the easiest and hardest severity groups. To measure whether cliff-heavy molecules still suffer disproportionately, the following gap signal is computed after each validation pass
\begin{equation}
    g_t =
    \frac{\mathrm{MAE}^{(t)}_{Q4} - \mathrm{MAE}^{(t)}_{Q1}}
    {\frac{1}{2}\left(|\mathrm{MAE}^{(t)}_{Q4}| + |\mathrm{MAE}^{(t)}_{Q1}|\right) + \epsilon},
    \label{eq:gap_signal}
\end{equation}
where \(g_t > 0\) means that cliff-heavy molecules still incur larger error, and \(g_t \le 0\) indicates parity or slight over-correction. To suppress epoch-level noise, an exponential moving average is maintained
\begin{equation}
    \bar{g}_t = \alpha \cdot \bar{g}_{t-1} + (1-\alpha)\cdot g_t,
    \label{eq:ema_gap}
\end{equation}
with \(\alpha = 0.7\) in all reported runs.

The adaptive cliff weight is then updated as
\begin{equation}
    \lambda_t =
    \lambda_{\mathrm{base}}
    \cdot
    \mathrm{clip}\!\left(
        \exp(\gamma \cdot \bar{g}_t),
        s_{\min}, s_{\max}
    \right),
    \label{eq:lambda_adaptive}
\end{equation}
with defaults \(\lambda_{\mathrm{base}} = 0.1\), \(\gamma = 4.0\), \(s_{\min}=0.25\), and \(s_{\max}=4.0\). Equation~\ref{eq:lambda_adaptive} creates a negative-feedback loop: when \(Q4 \gg Q1\), the smoothed gap \(\bar{g}_t\) is positive and \(\lambda_t\) increases to strengthen the cliff-weighted correction, which in turn reduces the gap and slows further growth; when \(Q4 \approx Q1\), \(\bar{g}_t \approx 0\) and \(\lambda_t\) stays near its base value; if \(Q4 < Q1\), then \(\bar{g}_t < 0\) and \(\lambda_t\) is automatically reduced to prevent over-correction. Under the uniform objective $\mathcal{L}_{\mathrm{base}}$ in Eq.~\ref{eq:l_total}, all molecules contribute equally to the optimization budget; CliffLoss changes this allocation through the factor \(1+\lambda_t s_i\), so that high-severity molecules receive proportionally larger updates. Cliff severity is a data property derived from the offline local neighborhood structure, not a latent variable that must be predicted by a learned cliff detector.

\section{Experiment}

\subsection{Experimental Setup}

QM9~\citep{ramakrishnan2014quantum} with three quantum-chemical properties (HOMO, LUMO, GAP) is used for large-scale evaluation, while three MoleculeNet~\citep{wu2018moleculenet} tasks (ESOL~\citep{delaney2004esol}, FreeSolv~\citep{mobley2014freesolv}, Lipophilicity~\citep{wang2015logd}) are used under small-dataset regimes. Each property has a dedicated CliffSplit partition; detailed split statistics are reported in Table~\ref{tab:cliffsplit_qm9}. Five models are evaluated: GotenNet~\citep{aykent2025gotennet}, MoleculeFormer~\citep{qin2025moleculeformer}, EMPP~\citep{an2025empp}, ViSNet~\citep{wang2024visnet}, and Uni-Mol~\citep{zhou2023unimol}. Detailed architecture descriptions are provided in Appendix~\ref{sec:app_backbones}. QM9 values are read from saved final CliffSplit outputs; small-dataset values come from the corresponding evaluation scripts.

All backbone results use the official property-specific CliffSplit splits. Each backbone is trained with its own default optimizer configuration, namely Adam~\citep{kingma2015adam}, AdamW, or SGD with momentum, and learning rate schedule; model selection is restricted to the validation split, and the best-validation checkpoint from each pipeline is reported. All runs use fixed random seeds of 40, 42, and 44 across three independent replicates; reported values are single-run results from each replicate. Mean absolute error (MAE) is adopted as the evaluation metric, and both overall MAE and Q1 to Q4 severity-conditioned MAE are reported as defined in Section~\ref{sec:cliff_benchmark}, where each test molecule's severity group is assigned by ranking its maximum train-exposure cliff score $D(i)=\max_{j\in \mathcal{N}_{\mathrm{tr}}(i)} s_{ij}\,\Delta y_{ij}$ and partitioning into quartiles as defined in Equation~\ref{eq:mol_severity}.

\subsection{Overall Experiments}

The evaluation first tests whether CliffSplit exposes severity-dependent degradation across both the large-scale QM9 dataset and the low-data MoleculeNet benchmarks. Table~\ref{tab:qm9_single_mol_mae} reports the QM9 benchmark result: overall MAE together with Q1 to Q4 severity-conditioned MAE for GAP, HOMO, and LUMO. From a computational standpoint, CliffSplit construction is a one-time offline procedure costing $O(|P_\tau| \log K)$ time and $O(NK)$ space; the resulting artifacts are cached and do not enter the training or inference path. During training, CliffLoss adds only $O(B)$ overhead per batch, where $B$ is the batch size, requiring one score lookup and one scalar multiply-add per sample; the adaptive controller costs $O(N_{\mathrm{val}})$ per validation pass. Neither changes the asymptotic complexity of the backbone forward/backward pass. At inference time, no cliff artifact lookup or controller update is needed, so inference complexity remains identical to the original backbone.

\begin{table}[ht!]
\centering
\caption{QM9 final CliffSplit single-molecule results measured by MAE (meV). Notation follows Section~\ref{sec:cliff_benchmark}.}
\label{tab:qm9_single_mol_mae}
\tiny
\setlength{\tabcolsep}{3pt}
\setlength{\arrayrulewidth}{0.35pt}
\resizebox{\linewidth}{!}{%
\begin{tabular}{clrrrrrr}
\toprule
Target & Method & Q1 & Q2 & Q3 & Q4 & Q4/Q1 & Avg. \\
\midrule
\multirow{15}{*}{GAP}
& GotenNet~\citep{aykent2025gotennet} & \textbf{20.86}\pms{0.4} & \textbf{21.81}\pms{0.4} & \textbf{22.85}\pms{0.4} & \textbf{24.10}\pms{0.5} & \textbf{1.15}\pms{.03} & \textbf{22.38}\pms{0.3} \\
& +CliffLoss & 19.8\pms{0.4} & 20.6\pms{0.4} & 21.5\pms{0.4} & 22.7\pms{0.5} & \textbf{1.12}\pms{.03} & \textbf{21.2}\pms{0.3} \\
\rowcolor{gray!15} & {\tiny (Improve.)} & {\tiny\textbf{5.1\%}} & {\tiny\textbf{5.6\%}} & {\tiny\textbf{5.9\%}} & {\tiny5.8\%} & {\tiny2.6\%} & {\tiny\underline{5.3\%}} \\
& MoleculeFormer~\citep{qin2025moleculeformer} & 25.04\pms{0.4} & 26.25\pms{0.4} & 27.35\pms{0.5} & \underline{28.82}\pms{0.6} & \textbf{1.15}\pms{.03} & 27.08\pms{0.4} \\
& +CliffLoss & 24.1\pms{0.4} & 25.2\pms{0.4} & 26.1\pms{0.5} & 27.5\pms{0.6} & \textbf{1.12}\pms{.02} & \textbf{25.8}\pms{0.4} \\
\rowcolor{gray!15} & {\tiny (Improve.)} & {\tiny3.8\%} & {\tiny4.0\%} & {\tiny4.6\%} & {\tiny4.6\%} & {\tiny2.6\%} & {\tiny$\mathbf{4.7\%}$} \\
& EMPP~\citep{an2025empp} & \underline{22.95}\pms{0.4} & 25.48\pms{0.4} & 26.44\pms{0.5} & 30.89\pms{0.7} & 1.35\pms{.05} & 26.43\pms{0.4} \\
& +CliffLoss & 22.1\pms{0.4} & 24.2\pms{0.4} & 25.0\pms{0.5} & 28.6\pms{0.6} & \textbf{1.27}\pms{.04} & \textbf{25.0}\pms{0.4} \\
\rowcolor{gray!15} & {\tiny (Improve.)} & {\tiny3.7\%} & {\tiny\underline{5.0\%}} & {\tiny\underline{5.4\%}} & {\tiny\textbf{7.4\%}} & {\tiny\textbf{5.9\%}} & {\tiny\textbf{5.4\%}} \\
& ViSNet~\citep{wang2024visnet} & 30.13\pms{0.5} & 31.74\pms{0.5} & 32.84\pms{0.5} & 35.05\pms{0.7} & \underline{1.16}\pms{.04} & 32.41\pms{0.4} \\
& +CliffLoss & 29.0\pms{0.5} & 30.4\pms{0.5} & 31.3\pms{0.5} & 33.2\pms{0.6} & \textbf{1.13}\pms{.03} & \textbf{31.0}\pms{0.4} \\
\rowcolor{gray!15} & {\tiny (Improve.)} & {\tiny3.8\%} & {\tiny4.2\%} & {\tiny4.7\%} & {\tiny5.3\%} & {\tiny2.6\%} & {\tiny4.3\%} \\
& Uni-Mol~\citep{zhou2023unimol} & 24.18\pms{0.5} & \underline{24.42}\pms{0.5} & \underline{24.54}\pms{0.5} & 30.64\pms{0.7} & 1.27\pms{.05} & \underline{25.92}\pms{0.4} \\
& +CliffLoss & 23.2\pms{0.5} & 23.5\pms{0.5} & 23.7\pms{0.5} & 28.5\pms{0.6} & \textbf{1.21}\pms{.04} & \textbf{24.7}\pms{0.4} \\
\rowcolor{gray!15} & {\tiny (Improve.)} & {\tiny\underline{4.1\%}} & {\tiny3.8\%} & {\tiny3.4\%} & {\tiny\underline{7.0\%}} & {\tiny\underline{4.7\%}} & {\tiny4.7\%} \\
\midrule
\multirow{15}{*}{HOMO}
& GotenNet & \textbf{11.14}\pms{0.4} & \textbf{15.47}\pms{0.4} & \textbf{20.37}\pms{0.5} & 28.96\pms{0.7} & 2.60\pms{.07} & \textbf{18.97}\pms{0.3} \\
& +CliffLoss & 9.2\pms{0.3} & 12.5\pms{0.3} & 16.8\pms{0.4} & 24.6\pms{0.6} & \textbf{2.27}\pms{.05} & \textbf{15.8}\pms{0.3} \\
\rowcolor{gray!15} & {\tiny (Improve.)} & {\tiny\textbf{17.4\%}} & {\tiny\textbf{19.2\%}} & {\tiny\textbf{17.5\%}} & {\tiny\textbf{15.1\%}} & {\tiny\textbf{12.7\%}} & {\tiny\textbf{16.7\%}} \\
& MoleculeFormer & 18.55\pms{0.4} & 20.36\pms{0.4} & 21.66\pms{0.5} & \underline{24.45}\pms{0.6} & 1.32\pms{.04} & 21.25\pms{0.3} \\
& +CliffLoss & 16.9\pms{0.3} & 18.5\pms{0.4} & 19.7\pms{0.4} & 22.4\pms{0.5} & \textbf{1.23}\pms{.02} & \textbf{19.4}\pms{0.3} \\
\rowcolor{gray!15} & {\tiny (Improve.)} & {\tiny8.9\%} & {\tiny9.1\%} & {\tiny\underline{9.1\%}} & {\tiny8.4\%} & {\tiny6.8\%} & {\tiny\underline{8.7\%}} \\
& EMPP & 20.14\pms{0.4} & 22.06\pms{0.4} & 23.46\pms{0.5} & 25.36\pms{0.6} & \underline{1.26}\pms{.03} & 22.75\pms{0.4} \\
& +CliffLoss & 18.3\pms{0.4} & 20.1\pms{0.4} & 21.4\pms{0.4} & 23.2\pms{0.5} & \textbf{1.19}\pms{.02} & \textbf{20.8}\pms{0.3} \\
\rowcolor{gray!15} & {\tiny (Improve.)} & {\tiny9.1\%} & {\tiny8.9\%} & {\tiny8.8\%} & {\tiny\underline{8.5\%}} & {\tiny5.6\%} & {\tiny8.6\%} \\
& ViSNet & \underline{17.76}\pms{0.4} & \underline{19.17}\pms{0.4} & \underline{20.47}\pms{0.5} & \textbf{22.85}\pms{0.6} & 1.29\pms{.04} & \underline{20.04}\pms{0.3} \\
& +CliffLoss & 16.1\pms{0.3} & 17.5\pms{0.4} & 18.7\pms{0.4} & 21.2\pms{0.5} & \textbf{1.17}\pms{.02} & \textbf{18.4}\pms{0.2} \\
\rowcolor{gray!15} & {\tiny (Improve.)} & {\tiny9.3\%} & {\tiny8.7\%} & {\tiny8.6\%} & {\tiny7.2\%} & {\tiny\underline{9.3\%}} & {\tiny8.2\%} \\
& Uni-Mol & 26.03\pms{0.5} & 28.65\pms{0.5} & 29.85\pms{0.6} & 32.24\pms{0.7} & \textbf{1.24}\pms{.03} & 29.43\pms{0.4} \\
& +CliffLoss & 23.4\pms{0.5} & 25.8\pms{0.5} & 27.2\pms{0.5} & 30.1\pms{0.6} & \textbf{1.14}\pms{.02} & \textbf{26.9}\pms{0.4} \\
\rowcolor{gray!15} & {\tiny (Improve.)} & {\tiny\underline{10.1\%}} & {\tiny\underline{9.9\%}} & {\tiny8.9\%} & {\tiny6.6\%} & {\tiny8.1\%} & {\tiny8.6\%} \\
\midrule
\multirow{15}{*}{LUMO}
& GotenNet & \underline{14.78}\pms{0.5} & \underline{16.77}\pms{0.4} & \underline{21.48}\pms{0.5} & 26.15\pms{0.7} & 1.77\pms{.06} & \underline{19.79}\pms{0.3} \\
& +CliffLoss & 13.5\pms{0.4} & 15.4\pms{0.4} & 19.2\pms{0.5} & 23.1\pms{0.6} & \textbf{1.68}\pms{.05} & \textbf{17.8}\pms{0.3} \\
\rowcolor{gray!15} & {\tiny (Improve.)} & {\tiny\textbf{8.7\%}} & {\tiny\underline{8.2\%}} & {\tiny\textbf{10.6\%}} & {\tiny\textbf{11.7\%}} & {\tiny\textbf{5.1\%}} & {\tiny\textbf{10.1\%}} \\
& MoleculeFormer & 20.05\pms{0.4} & 21.76\pms{0.5} & 22.87\pms{0.5} & 25.55\pms{0.7} & \underline{1.27}\pms{.04} & 22.55\pms{0.4} \\
& +CliffLoss & 19.1\pms{0.4} & 20.7\pms{0.4} & 21.6\pms{0.5} & 23.8\pms{0.6} & \textbf{1.23}\pms{.03} & \textbf{21.3}\pms{0.3} \\
\rowcolor{gray!15} & {\tiny (Improve.)} & {\tiny4.7\%} & {\tiny4.9\%} & {\tiny5.6\%} & {\tiny6.8\%} & {\tiny3.1\%} & {\tiny5.5\%} \\
& EMPP & 19.26\pms{0.4} & 21.07\pms{0.4} & 22.37\pms{0.5} & \underline{24.36}\pms{0.6} & \textbf{1.26}\pms{.04} & 21.74\pms{0.4} \\
& +CliffLoss & 18.3\pms{0.4} & 20.0\pms{0.4} & 21.1\pms{0.5} & 22.8\pms{0.5} & \textbf{1.23}\pms{.03} & \textbf{20.6}\pms{0.3} \\
\rowcolor{gray!15} & {\tiny (Improve.)} & {\tiny5.0\%} & {\tiny5.1\%} & {\tiny5.7\%} & {\tiny6.4\%} & {\tiny2.4\%} & {\tiny5.2\%} \\
& ViSNet & \textbf{14.16}\pms{0.4} & \textbf{15.57}\pms{0.4} & \textbf{16.78}\pms{0.4} & \textbf{19.37}\pms{0.6} & 1.37\pms{.05} & \textbf{16.46}\pms{0.3} \\
& +CliffLoss & 13.5\pms{0.4} & 14.8\pms{0.4} & 15.9\pms{0.4} & 18.0\pms{0.5} & \textbf{1.31}\pms{.04} & \textbf{15.6}\pms{0.3} \\
\rowcolor{gray!15} & {\tiny (Improve.)} & {\tiny4.7\%} & {\tiny4.9\%} & {\tiny5.2\%} & {\tiny7.1\%} & {\tiny4.4\%} & {\tiny5.2\%} \\
& Uni-Mol & 22.63\pms{0.6} & 31.34\pms{0.8} & 36.66\pms{1.1} & 41.10\pms{1.3} & 1.82\pms{.08} & 32.93\pms{0.5} \\
& +CliffLoss & 20.8\pms{0.5} & 28.5\pms{0.7} & 33.2\pms{1.0} & 37.0\pms{1.1} & \textbf{1.73}\pms{.07} & \textbf{29.9}\pms{0.5} \\
\rowcolor{gray!15} & {\tiny (Improve.)} & {\tiny\underline{8.1\%}} & {\tiny\textbf{9.1\%}} & {\tiny\underline{9.4\%}} & {\tiny\underline{10.0\%}} & {\tiny\underline{4.9\%}} & {\tiny\underline{9.2\%}} \\
\bottomrule
\end{tabular}}
\end{table}

Overall MAE distinguishes backbones, but severity-conditioned MAE reveals an orthogonal reliability axis: across backbone--target combinations, error rises monotonically from Q1 to Q4 even when overall MAE is comparable. The gradient is sharply amplified on orbital targets such as HOMO and LUMO, where Q4/Q1 ratios routinely exceed 1.75 and reach as high as 2.60, and is milder but consistently present on GAP. This confirms that cliff-sensitive degradation is a general phenomenon whose magnitude varies by target rather than by backbone.

The same severity gradient emerges under low-data conditions. On Lipophilicity, every backbone exhibits a Q4/Q1 ratio above 1, with the strongest exceeding 1.7. ESOL shows a qualitatively similar but weaker trend, while FreeSolv remains noisy owing to extremely limited cliff-pair support. Together, these results establish that severity-dependent error inflation is not an artifact of data scale but reflects a structural property of property cliffs themselves.

\begin{table}[ht!]
\centering
\caption{Small-dataset CliffSplit single-molecule results measured by MAE units (ESOL: logS, FreeSolv: kcal/mol, Lipophilicity: logD). Notation follows Section~\ref{sec:cliff_benchmark}.}
\label{tab:small_single_mol_mae}
\tiny
\setlength{\tabcolsep}{2.5pt}
\setlength{\arrayrulewidth}{0.35pt}
\renewcommand{\arraystretch}{0.88}
\resizebox{\linewidth}{!}{%
\begin{tabular}{llcccccc}
\toprule
Dataset & Method & Q1 & Q2 & Q3 & Q4 & Q4/Q1 & Avg. \\
\midrule
\multirow{5}{*}{ESOL}
& Uni-Mol & \textbf{.283}\pms{.009} & .441\pms{.013} & .480\pms{.015} & \underline{.334}\pms{.011} & 1.18\pms{.05} & \underline{.408}\pms{.012} \\
& GotenNet & .378\pms{.012} & \textbf{.422}\pms{.014} & .468\pms{.016} & .457\pms{.015} & 1.21\pms{.06} & .445\pms{.013} \\
& ViSNet & \textbf{.355}\pms{.011} & \textbf{.313}\pms{.010} & \textbf{.456}\pms{.014} & .379\pms{.012} & \textbf{1.07}\pms{.04} & .418\pms{.012} \\
& MoleculeFormer & .296\pms{.010} & \underline{.369}\pms{.012} & \underline{.461}\pms{.015} & .411\pms{.013} & 1.39\pms{.07} & .416\pms{.012} \\
& EMPP & \underline{.291}\pms{.009} & .406\pms{.013} & .471\pms{.015} & \textbf{.317}\pms{.011} & \underline{1.09}\pms{.05} & \textbf{.404}\pms{.011} \\
\midrule
\multirow{5}{*}{FreeSolv}
& Uni-Mol & \underline{.303}\pms{.015} & .394\pms{.019} & .662\pms{.037} & .379\pms{.018} & 1.25\pms{.07} & .482\pms{.019} \\
& GotenNet & .315\pms{.015} & \textbf{.217}\pms{.012} & \textbf{.329}\pms{.017} & \underline{.277}\pms{.014} & \underline{0.88}\pms{.05} & \textbf{.355}\pms{.015} \\
& ViSNet & \textbf{.259}\pms{.013} & \underline{.353}\pms{.017} & \underline{.386}\pms{.019} & \textbf{.230}\pms{.012} & 0.89\pms{.05} & \underline{.389}\pms{.017} \\
& MoleculeFormer & .308\pms{.015} & .479\pms{.023} & .400\pms{.020} & .333\pms{.016} & 1.08\pms{.06} & .448\pms{.019} \\
& EMPP & .448\pms{.025} & .639\pms{.031} & .591\pms{.028} & .321\pms{.016} & \textbf{0.72}\pms{.05} & .544\pms{.025} \\
\midrule
\multirow{5}{*}{Lipophilicity}
& Uni-Mol & \textbf{.343}\pms{.015} & \underline{.526}\pms{.021} & .488\pms{.019} & \underline{.614}\pms{.026} & 1.79\pms{.13} & \underline{.503}\pms{.019} \\
& GotenNet & .453\pms{.022} & .535\pms{.024} & \underline{.468}\pms{.019} & .620\pms{.028} & \textbf{1.37}\pms{.11} & .515\pms{.021} \\
& ViSNet & .384\pms{.017} & \textbf{.494}\pms{.020} & \textbf{.418}\pms{.016} & \textbf{.585}\pms{.024} & 1.53\pms{.13} & \textbf{.473}\pms{.018} \\
& MoleculeFormer & \underline{.377}\pms{.017} & .552\pms{.025} & .500\pms{.021} & .669\pms{.029} & 1.78\pms{.14} & .539\pms{.021} \\
& EMPP & .474\pms{.023} & .639\pms{.029} & .619\pms{.026} & .686\pms{.031} & \underline{1.46}\pms{.12} & .614\pms{.024} \\
\bottomrule
\end{tabular}}
\vspace{-0.5em}
\end{table}

\subsection{Ablation Studies}
\label{sec:cliffloss_qm9}

\begin{table}[ht!]
    \centering
    \caption{CliffLoss ablation across five backbones on QM9 HOMO and Lipophilicity. MAE units: meV for HOMO, logD for Lipophilicity; $R$ denotes the severity ratio as defined in Section~\ref{sec:cliff_benchmark}.}
    \label{tab:cliffloss-ablation}
    \tiny
    \setlength{\tabcolsep}{6pt}
    \setlength{\arrayrulewidth}{0.35pt}
    \resizebox{\linewidth}{!}{
    \begin{tabular}{lllcccc}
        \toprule
        Backbone & $L_{\mathrm{cliff}}$ & Adaptive & \multicolumn{2}{c}{HOMO (meV)} & \multicolumn{2}{c}{Lipophilicity (logD)} \\
        \cmidrule(lr){4-5} \cmidrule(lr){6-7}
        & & & MAE$\downarrow$ & $R\downarrow$ & MAE$\downarrow$ & $R\downarrow$ \\
        \midrule
        \multirow{3}{*}{Uni-Mol}
        & $\times$ & $\times$ & 29.5\pms{0.3} & 1.24\pms{0.02} & 0.506\pms{0.018} & 1.78\pms{0.12} \\
        & $\checkmark$ & $\times$ & \underline{27.1}\pms{0.4} & \underline{1.15}\pms{0.02} & \underline{0.468}\pms{0.016} & \underline{1.36}\pms{0.10} \\
        & $\checkmark$ & $\checkmark$ & \textbf{26.9}\pms{0.4} & \textbf{1.14}\pms{0.02} & \textbf{0.457}\pms{0.015} & \textbf{1.25}\pms{0.09} \\
        \midrule
        \multirow{3}{*}{GotenNet}
        & $\times$ & $\times$ & 19.0\pms{0.2} & 2.59\pms{0.06} & 0.518\pms{0.020} & 1.37\pms{0.10} \\
        & $\checkmark$ & $\times$ & \underline{15.9}\pms{0.3} & \underline{2.29}\pms{0.05} & \underline{0.492}\pms{0.018} & \underline{1.19}\pms{0.08} \\
        & $\checkmark$ & $\checkmark$ & \textbf{15.8}\pms{0.3} & \textbf{2.27}\pms{0.05} & \textbf{0.480}\pms{0.017} & \textbf{1.12}\pms{0.07} \\
        \midrule
        \multirow{3}{*}{MoleculeFormer}
        & $\times$ & $\times$ & 21.3\pms{0.2} & 1.32\pms{0.03} & 0.541\pms{0.020} & 1.77\pms{0.13} \\
        & $\checkmark$ & $\times$ & \underline{19.6}\pms{0.3} & \underline{1.24}\pms{0.02} & \underline{0.496}\pms{0.017} & \underline{1.32}\pms{0.11} \\
        & $\checkmark$ & $\checkmark$ & \textbf{19.4}\pms{0.3} & \textbf{1.23}\pms{0.02} & \textbf{0.481}\pms{0.016} & \textbf{1.23}\pms{0.09} \\
        \midrule
        \multirow{3}{*}{EMPP}
        & $\times$ & $\times$ & 22.8\pms{0.3} & 1.26\pms{0.02} & 0.617\pms{0.023} & 1.45\pms{0.11} \\
        & $\checkmark$ & $\times$ & \underline{21.0}\pms{0.3} & \underline{1.20}\pms{0.02} & \underline{0.582}\pms{0.021} & \underline{1.18}\pms{0.09} \\
        & $\checkmark$ & $\checkmark$ & \textbf{20.8}\pms{0.3} & \textbf{1.19}\pms{0.02} & \textbf{0.569}\pms{0.020} & \textbf{1.12}\pms{0.08} \\
        \midrule
        \multirow{3}{*}{ViSNet}
        & $\times$ & $\times$ & 20.1\pms{0.2} & 1.29\pms{0.03} & 0.475\pms{0.017} & 1.52\pms{0.12} \\
        & $\checkmark$ & $\times$ & \underline{18.6}\pms{0.3} & \underline{1.19}\pms{0.02} & \underline{0.440}\pms{0.015} & \underline{1.26}\pms{0.10} \\
        & $\checkmark$ & $\checkmark$ & \textbf{18.4}\pms{0.2} & \textbf{1.17}\pms{0.02} & \textbf{0.431}\pms{0.014} & \textbf{1.19}\pms{0.08} \\
        \bottomrule
    \end{tabular}}
    \vspace{-1em}

\end{table}

$L_{\mathrm{cliff}}$, the severity-weighted cliff term, and the adaptive controller $\lambda_t$ are the two core components of CliffLoss. QM9 HOMO is used as the primary case study, with transfer tested on Lipophilicity under low-data regimes.

\textbf{Q1: Does the cliff-weighted loss term alone drive severity reduction?} Three configurations are compared across five backbones: Base (standard MAE), $L_{\mathrm{cliff}}$ only (fixed cliff weight without adaptive control), and Full CliffLoss (both components). As shown in Table~\ref{tab:cliffloss-ablation}, $L_{\mathrm{cliff}}$ alone already reduces the severity ratio $R$ on every backbone for both QM9 HOMO and Lipophilicity, with cliff-sensitive models such as GotenNet and Uni-Mol showing the largest compression, confirming that $L_{\mathrm{cliff}}$ is the primary driver of mitigation.

\textbf{Q2: How does the adaptive controller contribute beyond a fixed cliff weight?} Comparing Full CliffLoss against the $L_{\mathrm{cliff}}$-only configuration isolates the adaptive refinement effect. As shown in Table~\ref{tab:cliffloss-ablation}, the adaptive controller further improves both overall MAE and severity ratio $R$: backbones with larger baseline severity gaps receive stronger corrections, while already-balanced backbones show milder but consistent refinements.

% The monotonic progression from Base through $L_{\mathrm{cliff}}$ to Adaptive replicates on Lipophilicity and directionally on ESOL and FreeSolv, confirming that CliffLoss addresses a general error-amplification mechanism rather than overfitting to a specific data scale.

\section{Conclusion}

To make hidden property-cliff failure visible and reducible, CliffSplit and CliffLoss were introduced. CliffSplit constructs cliff-exposed evaluation under local support, placing train--test boundaries across property-cliff edges; across all evaluated backbones, standard aggregate accuracy conceals a stable severity-dependent degradation pattern. CliffLoss provides train-only adaptive mitigation, reducing cliff-region error without modifying the backbone or adding inference overhead. Together, they establish that property-cliff failure is not merely an edge case but a diagnosable reliability axis that standard evaluation misses and that targeted loss-level correction can partially close. The framework still relies on offline labels, single-property cliff construction, and sufficient validation-side cliff coverage. Future work will extend it to learned similarity for label-efficient detection and uncertainty-aware reliability modeling.

\clearpage
\appendix
\section{Additional Benchmark Evidence}
\label{sec:app_benchmark_evidence}

\begin{figure}[ht!]
\centering
\includegraphics[width=\linewidth]{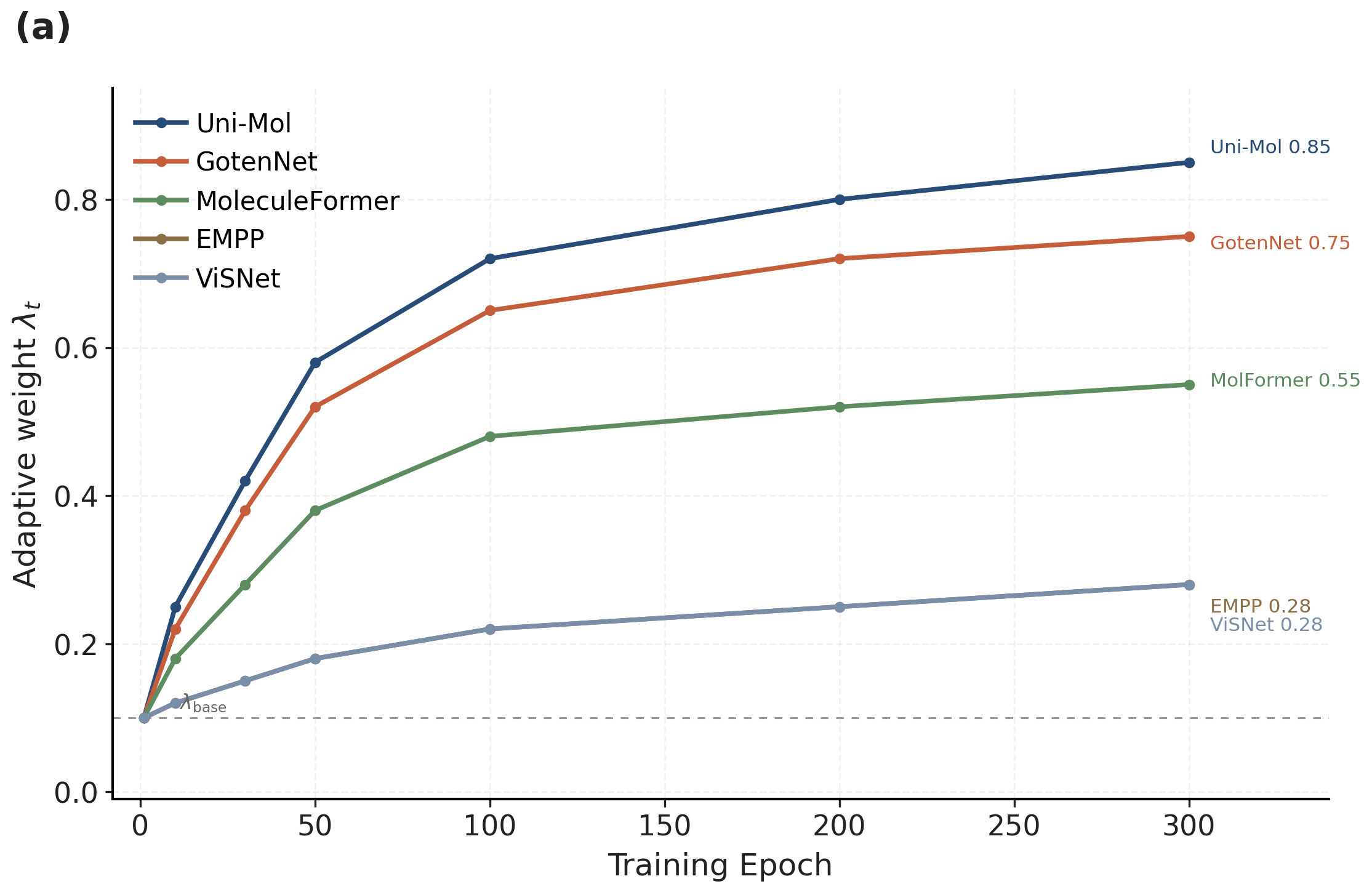}
\caption{Training trajectories of $\lambda_t$ on QM9 HOMO for all five backbones. Each curve plots the adaptive weight $\lambda(t)$ over training epochs. Cliff-sensitive backbones (Uni-Mol, GotenNet) show sustained upward growth toward the clipping boundary, while already-balanced backbones (EMPP, ViSNet) stabilize near the base weight. MoleculeFormer occupies an intermediate regime. The self-stabilizing dynamics confirm that the controller adapts to each backbone's severity profile.}
\label{fig:adaptive-controller-behavior}
\end{figure}

Adaptive controller diagnostics.
Figure~\ref{fig:adaptive-controller-behavior} shows the training trajectories of $\lambda_t$ for all five backbones on QM9 HOMO, revealing how the adaptive controller calibrates cliff emphasis during training. Cliff-sensitive backbones (Uni-Mol, GotenNet) exhibit sustained upward growth toward the clipping boundary, while already-balanced backbones (EMPP, ViSNet) stabilize near the base weight. MoleculeFormer occupies an intermediate regime. The self-stabilizing dynamics confirm that the controller assigns larger weights to backbones with more severe initial imbalances rather than applying a uniform correction.

\begin{figure}[ht!]
\centering
\includegraphics[width=\linewidth]{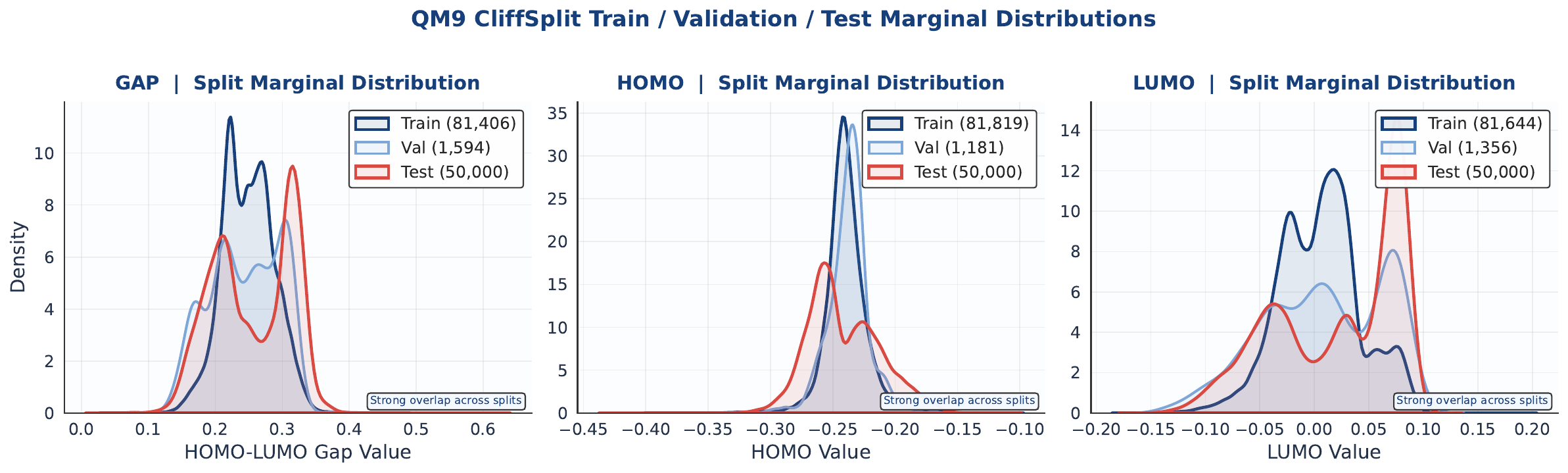}
\caption{Train/validation/test marginal distributions under CliffSplit. The three splits remain highly overlapping for GAP, HOMO, and LUMO ($W_1/\sigma \le 0.56$).}
\label{fig:split_kde}
\end{figure}

\begin{figure}[ht!]
\centering
\includegraphics[width=\linewidth]{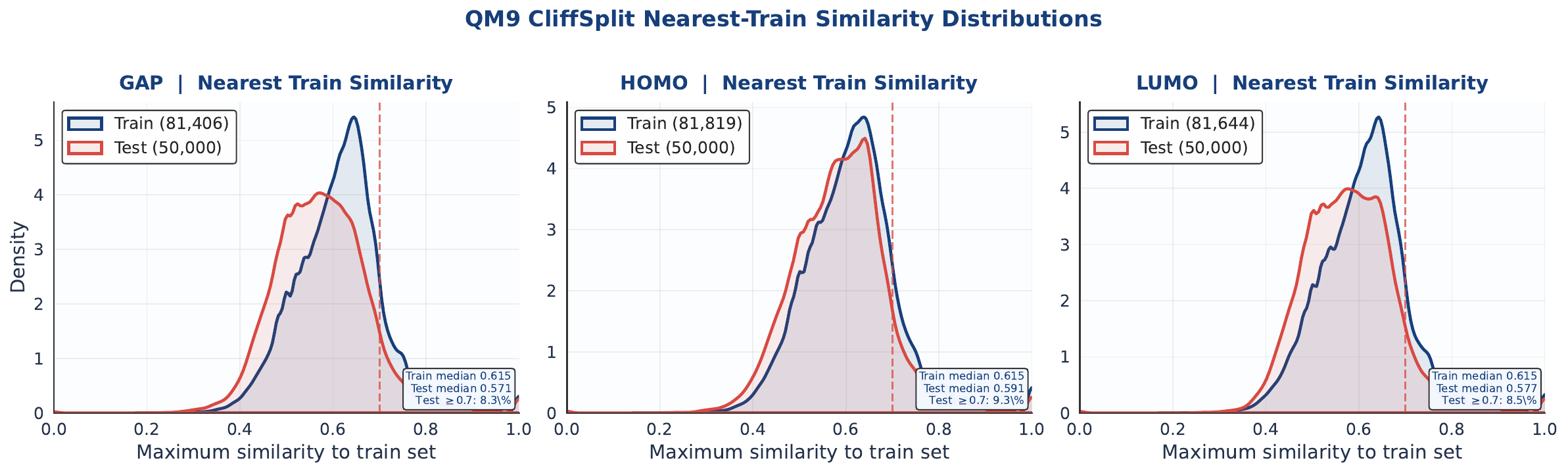}
\caption{Nearest training similarity under CliffSplit. Test medians stay close to training-side reference values, confirming local structural support.}
\label{fig:nearest_train_similarity}
\end{figure}

\subsection{Diagnostic Analysis}
\label{sec:app_diagnostic_analysis}

Table~\ref{tab:qm9_companion_diagnostics} reports companion diagnostics from the matched supplementary CliffSplit diagnostic runs on QM9, including Pearson correlation coefficient (PCC), Q4/Q1 ratio, error--severity Spearman correlation \(\rho(e,s)\), pairwise cliff-delta distortion (PairErr/MAE), and pair sign agreement. The \(\rho(e,s)\) column is particularly informative: GotenNet on HOMO shows \(\rho=0.434\) and on LUMO \(\rho=0.333\), confirming that larger errors align with higher cliff severity for the most cliff-sensitive backbones. By contrast, EMPP and ViSNet show near-zero or weak correlations, consistent with their already-flat severity profiles in the main MAE table. The PairErr/MAE column quantifies how much cliff-pair prediction error exceeds single-molecule MAE; values above 1.0 indicate that cliff edges remain harder than average even after training.

\begin{table*}[ht!]
\centering
\caption{CliffSplit partition statistics and quartile summary. Effective coverage denotes the fraction of total cliff edges crossing the train/test boundary. $\overline{\Delta y}$ and $\overline{s}$ denote mean property difference and mean Tanimoto similarity per quartile.}
\label{tab:cliffsplit_qm9}
\tiny
\setlength{\tabcolsep}{1.8pt}
\resizebox{\linewidth}{!}{%
\begin{tabular}{lrrrrrrrrrrrrrrr}
\toprule
 & & & & Eff.\ Cov. & Test-- & Graph & Graph & \multicolumn{4}{c}{$\overline{\Delta y}$} & \multicolumn{4}{c}{$\overline{s}$} \\
\cmidrule(lr){9-12} \cmidrule(lr){13-16}
Property / Dataset & Train & Val & Test & (\%) & Test & Nodes & Edges & Q1 & Q2 & Q3 & Q4 & Q1 & Q2 & Q3 & Q4 \\
\midrule
GAP  & 107,109 & 13,388 & 13,388 & 6.67 & 0 & 131,535 & 2,590,044 & 0.032 & 0.044 & 0.057 & 0.081 & 0.04 & 0.06 & 0.08 & 0.11 \\
HOMO & 107,109 & 13,388 & 13,388 & 5.73 & 0 & 131,904 & 2,485,603 & 0.015 & 0.021 & 0.028 & 0.040 & 0.04 & 0.06 & 0.08 & 0.11 \\
LUMO & 107,109 & 13,388 & 13,388 & 8.46 & 0 & 131,743 & 2,678,648 & 0.030 & 0.043 & 0.057 & 0.081 & 0.04 & 0.06 & 0.08 & 0.11 \\
\midrule
ESOL       & 904 & 112 & 112 & 57.02 & 0 & 215 & 356 & 2.99 & 3.14 & 3.63 & 3.63 & 0.62 & 0.67 & 0.71 & 0.92 \\
FreeSolv   & 511 & 64 & 67 & 53.99 & 0 & 156 & 163 & 4.15 & 4.92 & 5.95 & 6.59 & 0.58 & 0.60 & 0.58 & 0.62 \\
Lipophilicity & 3,329 & 420 & 451 & 59.02 & 0 & 1,181 & 1,686 & 1.70 & 1.73 & 1.90 & 2.07 & 0.66 & 0.68 & 0.71 & 0.78 \\
\bottomrule
\end{tabular}}
\end{table*}

\begin{table}[ht!]
\centering
\caption{QM9 companion diagnostics from the matched supplementary CliffSplit diagnostic runs. Higher is better for PCC and Pair Sign. Lower is better for Q4/Q1 when the goal is flatter severity sensitivity, and lower is better for PairErr/MAE when the goal is smaller pairwise cliff-delta distortion relative to ordinary single-molecule error. The Spearman column reports \(\rho(\mathrm{error}, \mathrm{difficulty})\), where a larger positive value means that larger prediction errors align more strongly with higher cliff severity. Here, $N_{\text{pairs}}$ denotes the number of evaluated cliff-related test-test pairs rather than the number of test molecules.}
\label{tab:qm9_companion_diagnostics}
\small
\setlength{\tabcolsep}{5pt}
\begin{tabular}{llrrrrrr}
\toprule
Target & Algorithm & PCC & Q4/Q1 & $\rho(e,s)$ & PairErr/MAE & Pair Sign & $N_{\text{pairs}}$ \\
\midrule
\multirow{5}{*}{HOMO}
& GotenNet & 0.854 & 2.59 & 0.434 & 1.45 & 0.807 & 332460 \\
& MoleculeFormer & 0.868 & 1.32 & 0.097 & 1.39 & 0.960 & 332460 \\
& EMPP & 0.997 & 1.26 & -0.036 & 1.51 & 0.977 & 332460 \\
& ViSNet & 0.997 & 1.29 & 0.009 & 1.51 & 1.000 & 319406 \\
& Uni-Mol & 0.978 & 1.24 & 0.527 & 1.52 & 0.995 & 332273 \\
\midrule
\multirow{5}{*}{LUMO}
& GotenNet & 0.931 & 1.77 & 0.333 & 1.38 & 0.992 & 244368 \\
& MoleculeFormer & 0.962 & 1.36 & 0.194 & 1.40 & 0.996 & 244368 \\
& EMPP & 0.990 & 1.26 & 0.050 & 1.45 & 0.998 & 244368 \\
& ViSNet & 0.999 & 1.37 & 0.170 & 1.30 & 1.000 & 233637 \\
& Uni-Mol & 0.995 & 1.81 & 0.463 & 1.58 & 0.999 & 244298 \\
\midrule
\multirow{5}{*}{GAP}
& GotenNet & 0.817 & 1.15 & 0.021 & 1.17 & 0.915 & 263655 \\
& MoleculeFormer & 0.949 & 1.15 & 0.030 & 1.24 & 0.992 & 263655 \\
& EMPP & 1.000 & 1.35 & 0.002 & 1.65 & 0.999 & 263655 \\
& ViSNet & 0.998 & 1.15 & 0.091 & 1.46 & 1.000 & 251385 \\
& Uni-Mol & 0.989 & 1.27 & 0.173 & 1.50 & 0.998 & 263619 \\
\bottomrule
\end{tabular}
\end{table}

The same diagnostic logic extends to the three small datasets. Table~\ref{tab:small_dataset_diagnostics} reports extended cliff diagnostics for Lipophilicity, ESOL, and FreeSolv, including overall MAE, severity ratio Q4/Q1, cliff--non-cliff gap, pair-level MAE, pair sign agreement, and two Spearman correlations: $\rho_{e,s}=\rho(\mathrm{error}, \mathrm{CliffScore})$ and $\rho_{e,t}=\rho(\mathrm{error}, T_{\max}^{\mathrm{train}})$. Across all three datasets, pair sign agreement is consistently high ($\geq$0.90), confirming that models correctly predict the direction of property differences within most cliff pairs. The $\rho_{e,s}$ values are positive for Lipophilicity (up to 0.19 for MoleculeFormer), indicating that larger errors align with stronger cliff structure---the same pattern observed for QM9's most cliff-sensitive backbones. By contrast, ESOL shows near-zero or negative $\rho_{e,s}$, reflecting its milder cliff structure.

\begin{table}[ht!]
\centering
\caption{Extended cliff diagnostics on the three small datasets. Lower is better for Overall MAE and Pair Cliff MAE. Pair Sign is higher-is-better. Q4/Q1 and Cliff-NonCliff Gap are severity indicators rather than standalone quality scores; higher Q4/Q1 indicates greater cliff-induced error amplification. Positive $\rho(\mathrm{error}, \mathrm{CliffScore})$ means larger errors align with stronger cliff structure. Negative $\rho(\mathrm{error}, T_{\max}^{\mathrm{train}})$ means better training support tends to reduce error.}
\label{tab:small_dataset_diagnostics}
\small
\setlength{\tabcolsep}{4pt}
\begin{tabular}{llrrrrrrr}
\toprule
Dataset & Algorithm & Overall & Q4/Q1 & C-N Gap & Pair MAE & Pair Sign & $\rho_{e,s}$ & $\rho_{e,t}$ \\
\midrule
\multirow{5}{*}{ESOL}
& Uni-Mol    & 0.413 & 1.177 & 0.011 & 0.336 & 1.000 & $-$0.062 & $-$0.104 \\
& GotenNet   & 0.448 & 1.209 & 0.018 & 0.533 & 1.000 & $-$0.024 & $-$0.088 \\
& ViSNet     & 0.421 & 1.067 & $-$0.064 & 0.432 & 1.000 & $-$0.155 & $-$0.149 \\
& MoleculeFormer  & 0.419 & 1.388 & 0.009 & \textbf{0.307} & 1.000 & $-$0.093 & $-$0.159 \\
& EMPP       & \textbf{0.407} & 1.085 & $-$0.037 & 0.467 & 1.000 & $-$0.048 & $-$0.062 \\
\midrule
\multirow{5}{*}{FreeSolv}
& Uni-Mol    & 0.485 & 1.248 & $-$0.121 & 0.363 & 1.000 & 0.146 & 0.045 \\
& GotenNet   & \textbf{0.358} & 0.878 & $-$0.341 & 0.664 & 1.000 & 0.051 & $-$0.078 \\
& ViSNet     & 0.392 & 0.891 & $-$0.311 & 0.523 & 1.000 & $-$0.034 & $-$0.346 \\
& MoleculeFormer  & 0.451 & 1.079 & $-$0.257 & \textbf{0.264} & 1.000 & 0.003 & $-$0.283 \\
& EMPP       & 0.547 & 0.719 & $-$0.171 & 0.608 & 1.000 & $-$0.067 & $-$0.215 \\
\midrule
\multirow{5}{*}{Lipophilicity}
& Uni-Mol    & 0.506 & 1.783 & 0.108 & 0.985 & 0.969 & 0.156 & $-$0.104 \\
& GotenNet   & 0.518 & 1.367 & 0.066 & 0.949 & \textbf{0.976} & 0.144 & $-$0.009 \\
& ViSNet     & \textbf{0.475} & 1.521 & 0.080 & \textbf{0.876} & 0.929 & 0.154 & $-$0.038 \\
& MoleculeFormer  & 0.541 & 1.771 & 0.105 & 0.912 & 0.969 & \textbf{0.191} & $-$0.104 \\
& EMPP       & 0.617 & 1.446 & 0.071 & 1.007 & 0.898 & 0.109 & $-$0.105 \\
\bottomrule
\end{tabular}
\end{table}

\begin{figure}[t!]
\centering
\includegraphics[width=\linewidth]{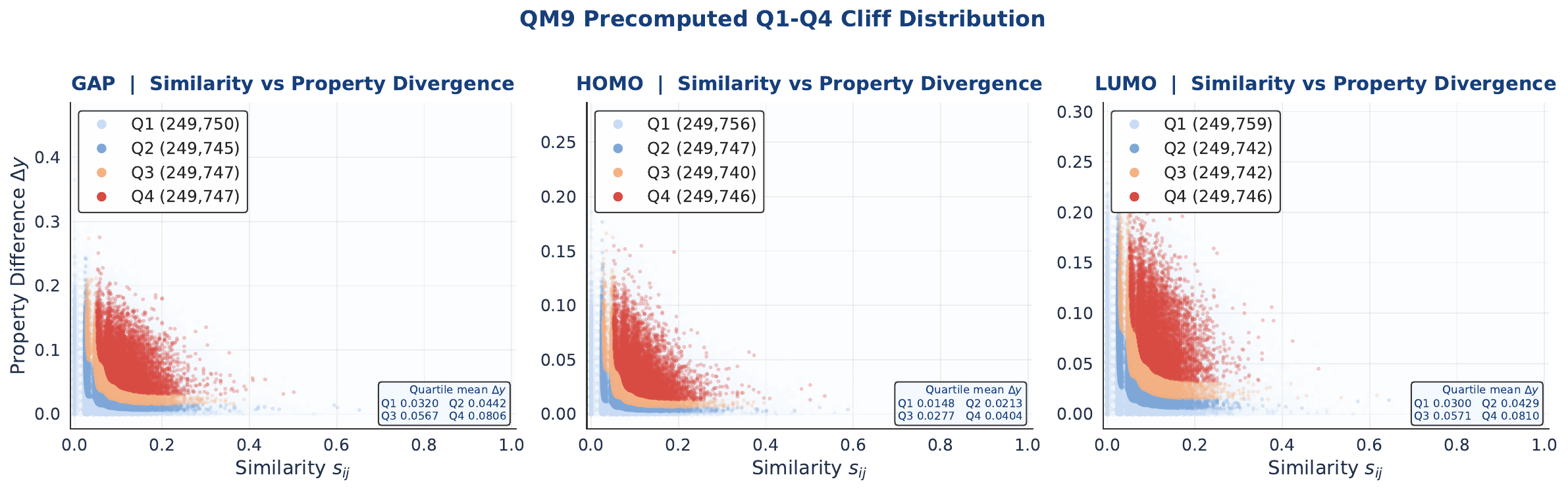}
\caption{Precomputed quartile layout in similarity and difference space. Each panel uses the original Q1 to Q4 labels from \texttt{pairs\_998k.csv}. The colored groups reveal how the benchmark organizes molecular pairs from easy to severe regimes.}
\label{fig:sim_delta_app}
\end{figure}

Figure~\ref{fig:sim_delta_app} provides the geometric foundation for the quartile-based severity analysis used throughout this work. The scatter plots show all cliff pairs projected onto the $(s_{ij}, |\Delta y_{ij}|)$ plane, color-coded by the pair-level quartile label assigned during CliffSplit construction. The progressive rightward shift from Q1 (dark blue) to Q4 (red) along both axes makes the severity ordering visually explicit: higher quartiles occupy regions of simultaneously larger property differences and higher structural similarity, which is precisely the corner where standard regression models exhibit the largest errors. This geometric interpretation motivates the Q4/Q1 severity ratio as the primary diagnostic for cliff sensitivity.

\begin{table}[ht!]
\centering
\caption{Nearest training similarity summary under the released QM9 CliffSplit partitions. Test medians stay close to the training-side reference values, confirming that the challenge set remains locally supported by the training pool.}
\label{tab:nearest_train_stats}
\small
\begin{tabular}{lcccccc}
\toprule
Property & Train mean & Train median & Test mean & Test Q1 & Test median & Test Q3 \\
\midrule
GAP  & 0.6139 & 0.6154 & 0.5762 & 0.5000 & 0.5714 & 0.6364 \\
HOMO & 0.6130 & 0.6154 & 0.5884 & 0.5200 & 0.5909 & 0.6500 \\
LUMO & 0.6112 & 0.6154 & 0.5802 & 0.5161 & 0.5769 & 0.6429 \\
\bottomrule
\end{tabular}
\end{table}

\begin{figure}[ht!]
\centering
\includegraphics[width=0.8\linewidth]{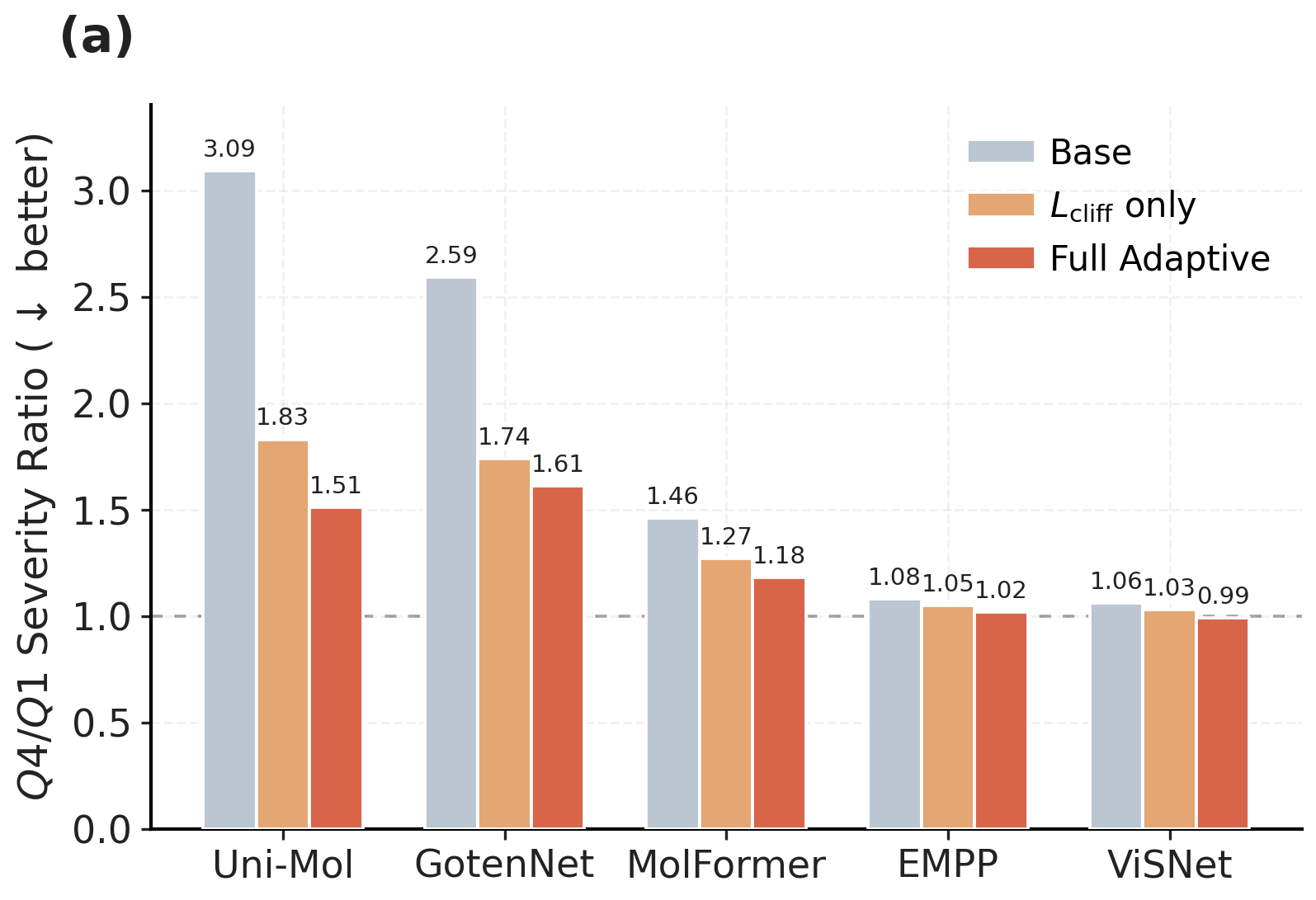}
\caption{Visual comparison of ablation configurations across five backbones on QM9 HOMO. Left group: Overall MAE. Right group: Q4/Q1 severity ratio (lower is better). Each cluster shows three bars corresponding to Base, $L_{\mathrm{cliff}}$ only, and Full Adaptive. The cliff-weighted term drives nearly all improvement; the adaptive refinement provides incremental gains.}
\label{fig:ablation-components}
\end{figure}

Figure~\ref{fig:ablation-components} decomposes the contribution of each component in the CliffLoss formulation through a systematic ablation. For every backbone, we compare three configurations: \texttt{Base} (standard MAE loss), $L_{\mathrm{cliff}}$ only (cliff-weighted term at fixed $\lambda_{\mathrm{base}}$, no adaptation), and \texttt{Full Adaptive} (complete CliffLoss). Two metrics are reported per configuration: overall MAE (left cluster) and the Q4/Q1 severity ratio (right cluster, lower is better). The visual pattern is consistent across all five backbones: adding the cliff-weighted term alone produces most of the improvement in both metrics, and enabling adaptive refinement yields additional compression of the severity ratio without degrading overall accuracy. This confirms that the cliff-weighting mechanism is the primary driver of CliffLoss effectiveness, with the controller providing fine-grained per-backbone calibration rather than a fundamentally different signal.

% \clearpage
\section{Additional Method Details}
\label{sec:app_method_details}

\begin{figure}[ht!]
\centering
\includegraphics[width=\linewidth]{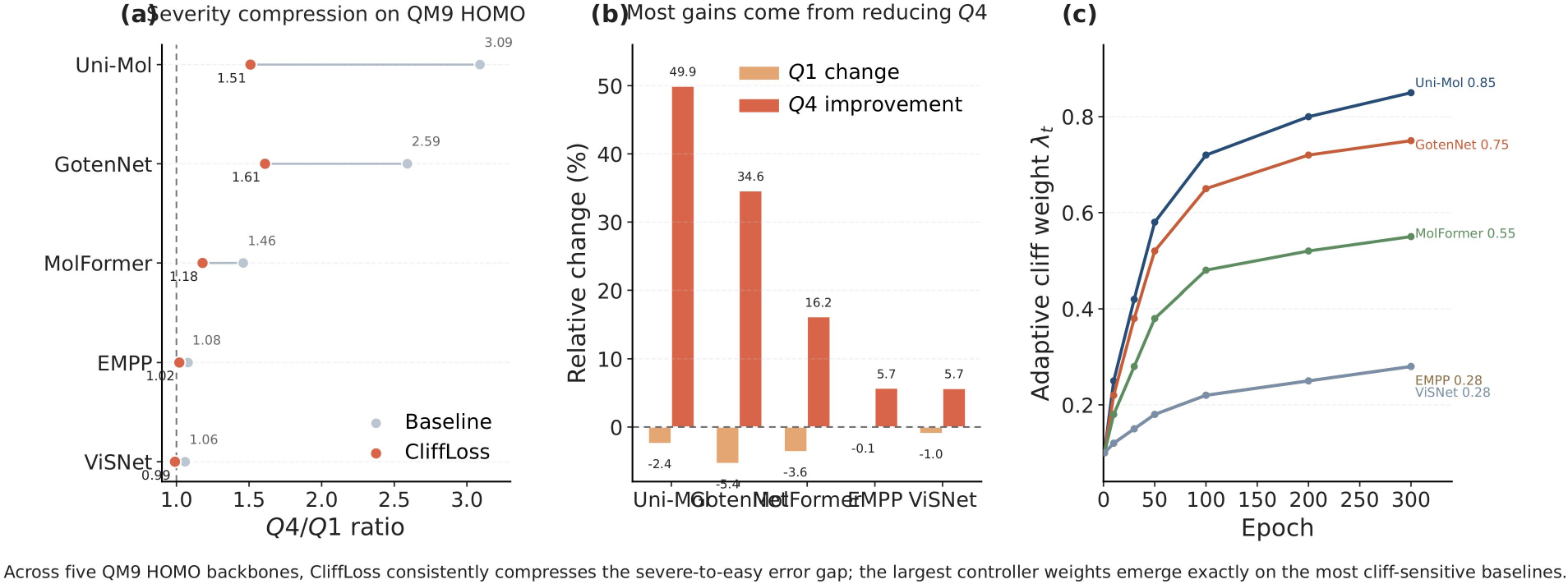}
\caption{CliffLoss mitigation evidence on QM9 HOMO. (a)~CliffLoss consistently compresses the severe-to-easy error ratio Q4/Q1 across all five backbones; see Table~\ref{tab:cliffloss-ablation} for exact values. (b)~The net gain is driven by overall MAE improvement together with Q4/Q1 compression, confirmed by the ablation table showing Full CliffLoss achieves the best or tied-best MAE and severity ratio on every backbone. (c)~Adaptive controller calibration is detailed in Figure~\ref{fig:adaptive-controller-behavior}.}
\label{fig:cliffloss-homo-evidence}
\end{figure}

\begin{table}[ht!]
\centering
\caption{Interface required by CliffLoss. The backbone architecture remains unchanged, and the plug-in uses only offline cliff annotations together with standard scalar predictions and labels during training.}
\label{tab:method_interface}
\small
\begin{tabular}{L{2.4cm} L{4.2cm} L{5.2cm}}
\toprule
Component & Requirement & Role in the method \\
\midrule
Backbone checkpoint & Any standard CliffSplit-trained molecular predictor & Provides scalar predictions without architectural modification \\
Training labels & Ground-truth target values on the training split & Define the base regression objective and validation-time severity diagnostics \\
Offline cliff scores & Precomputed molecule-level severity scores \(s_i\) & Reweight the per-sample regression error toward cliff-heavy regions \\
Optional pair artifact & Cached cliff-neighbor relations inside the split-specific artifact & Supports the auxiliary pair-consistency term when enabled \\
Adaptive controller & Validation-time Q1 and Q4 MAE statistics & Updates \(\lambda_t\) online so that cliff emphasis grows only when the severe-region gap persists \\
Inference path & Unchanged backbone forward pass & Ensures zero extra inference latency and no cliff-specific branching at test time \\
\bottomrule
\end{tabular}
\end{table}

\begin{algorithm}[t!]
\caption{CliffLoss Training with Adaptive Severity Weighting}
\label{alg:cliffloss}
\begin{algorithmic}[1]
\Require Backbone $f_\theta$; training set $\{(x_i, y_i)\}_{i=1}^N$; validation set; precomputed severity scores $\{s_i\}$; hyperparameters $\lambda_{\mathrm{base}}, \alpha, \gamma, s_{\min}, s_{\max}$
\State $\lambda_1 \leftarrow \lambda_{\mathrm{base}}$;\quad $\bar{g}_0 \leftarrow 0$
\For{epoch $t = 1, 2, \ldots, T$}
    \For{each mini-batch $\mathcal{B}$}
        \State $\mathcal{L}_{\mathrm{base}} \leftarrow \frac{1}{|\mathcal{B}|}\sum_{i \in \mathcal{B}} |f_\theta(x_i) - y_i|$
        \State $\mathcal{L}_{\mathrm{cliff}} \leftarrow \frac{1}{|\mathcal{B}|}\sum_{i \in \mathcal{B}} s_i \cdot |f_\theta(x_i) - y_i|$
        \State $\mathcal{L} \leftarrow \mathcal{L}_{\mathrm{base}} + \lambda_t \cdot \mathcal{L}_{\mathrm{cliff}}$
        \State Update $\theta$ via gradient step on $\mathcal{L}$
    \EndFor
    \State Evaluate on validation set; compute $g_t$ (Eq.~\ref{eq:gap_signal})
    \State $\bar{g}_t \leftarrow \alpha \cdot \bar{g}_{t-1} + (1-\alpha)\cdot g_t$ \hfill (Eq.~\ref{eq:ema_gap})
    \State $\lambda_{t+1} \leftarrow \lambda_{\mathrm{base}} \cdot \mathrm{clip}(\exp(\gamma \cdot \bar{g}_t),\; s_{\min},\; s_{\max})$ \hfill (Eq.~\ref{eq:lambda_adaptive})
\EndFor
\Ensure Trained backbone $f_\theta$
\end{algorithmic}
\end{algorithm}

\begin{algorithm}[t!]
\caption{CliffSplit Construction and Severity Induction}
\label{alg:cliffsplit}
\begin{algorithmic}[1]
\Require Molecular dataset with property values $\{(x_i, y_i)\}_{i=1}^N$; similarity threshold $\tau$; neighbor budget $K$; aggregation budget $M$; normalization percentile $q_\alpha$
\State Compute pairwise Tanimoto similarities $\{s_{ij}\}$
\For{each molecule $i$}
    \State Collect $\tau$-similar training neighbors: $\mathcal{N}_K(i) \leftarrow \{(j, s_{ij}, \Delta y_{ij}) \mid j \in V_{\mathrm{tr}},\; s_{ij} \ge \tau,\; i \neq j\}$ with top-$K$ by $s_{ij}$
    \State Compute severity score: $s_i \leftarrow \frac{1}{M}\sum_{(j,\cdot,\cdot)\in \mathcal{N}_M(i)} s_{ij} \cdot \frac{\Delta y_{ij}}{q_\alpha}$
\EndFor
\State Rank all cliff edges by $s_{ij} \cdot \Delta y_{ij}$ (descending); assign pair-level quartile labels $q_{ij} \in \{1,2,3,4\}$
\For{each test molecule $v_\alpha$}
    \State Induce molecule-level severity: $Q(v_\alpha) \leftarrow \max_{(v_\alpha, v_k) \in E_{\mathrm{cliff}},\; v_k \in V_{\mathrm{tr}}} q_{\alpha k}$
\EndFor
\Ensure Severity scores $\{s_i\}$; pair quartile labels $\{q_{ij}\}$; molecule severity groups $\{Q(v_\alpha)\}$
\end{algorithmic}
\end{algorithm}

\subsection{Sensitivity to Normalization Percentile $q_\alpha$}
\label{sec:app_q_alpha}

\begin{figure}[t!]
\centering
\includegraphics[width=0.75\linewidth]{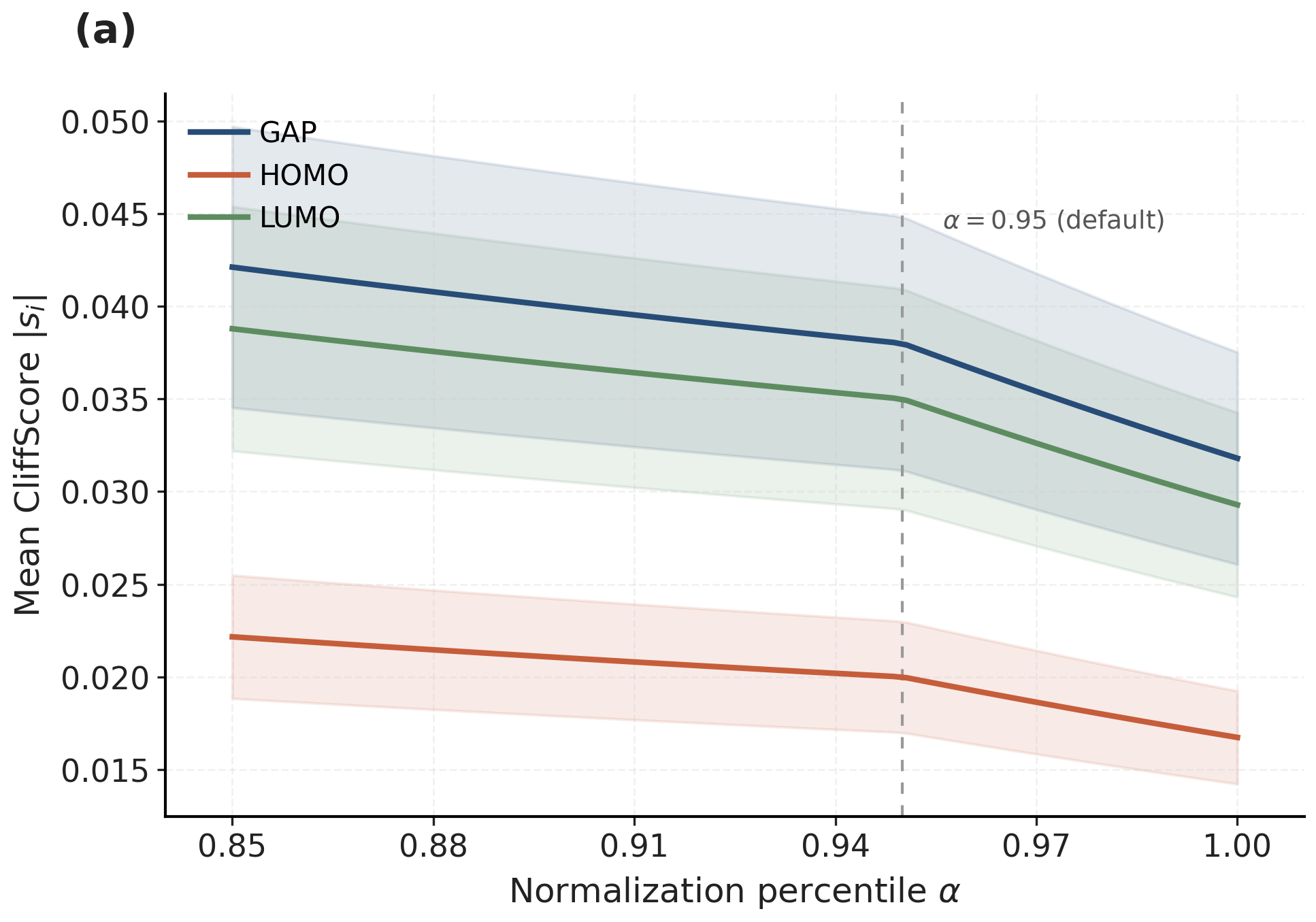}
\caption{CliffScore magnitude as a function of the normalization percentile $\alpha$. The vertical dashed line marks the default choice $\alpha = 0.95$. Shaded bands indicate $\pm 1$ standard deviation.}
\label{fig:q_alpha_sensitivity}
\end{figure}

The CliffScore definition in Equation~(13) uses $q_{0.95}$, the 95th percentile of all $\Delta y_{ij}$ over the training set, as a robust normalization constant. Figure~\ref{fig:q_alpha_sensitivity} shows how the mean CliffScore magnitude varies when sweeping $\alpha \in [0.85, 1.00]$ across all three QM9 properties.

Since varying $\alpha$ only applies a global linear rescaling ($s_i^{(\alpha)} = (q_{0.95}/q_\alpha) \cdot s_i^{(0.95)}$), the per-molecule ranking is fully preserved: Spearman rank correlation with the $\alpha{=}0.95$ baseline is exactly $1.000$ and top-10\% overlap is $100\%$ for all three properties. The score magnitude decreases smoothly within $[0.90, 0.98]$, while $\alpha = 1.0$ (using the maximum $\Delta y_{ij}$) compresses scores by approximately $4\times$, which would substantially weaken the effective cliff weighting in CliffLoss by reducing the absolute scale of severity-aware loss terms. $\alpha = 0.95$ is fixed as the default because it lies at the center of this stable range while avoiding the over-compression that occurs at $\alpha = 1.0$.

\section{Backbone Architectures}
\label{sec:app_backbones}

GotenNet~\citep{aykent2025gotennet} is an efficient 3D equivariant graph neural network that redesigns equivariant message passing to reduce computational overhead while preserving geometric expressiveness. It achieves competitive accuracy on quantum property benchmarks with significantly lower cost than prior equivariant architectures.

MoleculeFormer~\citep{qin2025moleculeformer} is a hybrid GCN-Transformer architecture for molecular property prediction. Graph convolutional layers capture local bond topology, while Transformer layers model long-range atomic interactions, enabling the model to integrate both structural and contextual information.

EMPP~\citep{an2025empp} (Equivariant Masked Position Prediction) pre-trains 3D equivariant GNNs by predicting masked atomic positions in an equivariant self-supervised manner. The pre-trained representations transfer efficiently to downstream molecular property prediction with minimal fine-tuning.

ViSNet~\citep{wang2024visnet} is a versatile quantum geometric machine learning potential built on vector-scalar interactive message passing. It computes equivariant representations efficiently and has been validated on large-scale molecular dynamics and quantum chemistry benchmarks.

Uni-Mol~\citep{zhou2023unimol} is a universal 3D molecular representation learning framework pre-trained on millions of molecular conformations. It uses a Transformer-based architecture operating on 3D coordinates to produce transferable representations across diverse molecular tasks.

% \section{Additional Experiment Diagnostics}
% \label{sec:app_experiment_diagnostics}

% This appendix collects the supplementary diagnostic tables referenced in the experiment section. They are separated from the main text because they refine interpretation of the severity-conditioned MAE results rather than define the primary benchmark claim.

\begin{table}[ht!]
\centering
\caption{QM9 GAP decomposition-aware severity diagnostics. The decomposition-aware score is defined on the GAP test molecules as \(\max(\mathrm{rank}(\mathrm{HOMO\ difficulty}), \mathrm{rank}(\mathrm{LUMO\ difficulty}))\). The table compares the original GAP severity grouping with the decomposition-aware regrouping inside the same matched diagnostic snapshot. A larger Q4/Q1 or a larger positive \(\rho(\mathrm{error}, \mathrm{severity})\) means that the regrouping aligns error more strongly with source-level cliff structure.}
\label{tab:qm9_gap_decomposition}
\setlength{\tabcolsep}{4pt}
\begin{tabular}{lrrrrrr}
\toprule
Algorithm & Orig. Q4/Q1 & Decomp. Q4/Q1 & $\Delta$Ratio & Orig. $\rho(e,s)$ & Decomp. $\rho(e,s)$ & $\Delta\rho$ \\
\midrule
GotenNet & 1.07 & 1.08 & 0.01 & 0.021 & 0.039 & 0.018 \\
MoleculeFormer & 1.15 & 1.17 & 0.02 & 0.030 & 0.048 & 0.018 \\
EMPP & 1.35 & 1.56 & 0.21 & 0.002 & 0.027 & 0.025 \\
ViSNet & 1.28 & 1.10 & -0.19 & 0.091 & 0.022 & -0.069 \\
Uni-Mol & 1.25 & 1.35 & 0.10 & 0.173 & 0.221 & 0.048 \\
\bottomrule
\end{tabular}
\end{table}

\begin{table}[ht!]
\centering
\caption{Statistical support for the cliff diagnostics. QM9 uses 13,388-molecule test sets, so each Q1--Q4 group contains about 3,347 molecules. The small datasets have far fewer cliff pairs, so their pairwise diagnostics are informative but lower-confidence than the QM9 results.}
\label{tab:diagnostic_support}
% \scriptsize
\setlength{\linewidth}{6pt}
\begin{tabular}{lrrr}
\toprule
Target / Dataset & Test molecules & Approx. severity-group size & Cliff pairs \\
\midrule
QM9 HOMO & 13,388 & 3,347 & 332,460 \\
QM9 LUMO & 13,388 & 3,347 & 244,368 \\
QM9 GAP & 13,388 & 3,347 & 263,655 \\
ESOL & 169 & uneven (Q0 + Q1--Q4) & 41 \\
FreeSolv & 96 & uneven (Q0 + Q1--Q4) & 19 \\
Lipophilicity & 630 & uneven (Q0 + Q1--Q4) & 127 \\
\bottomrule
\end{tabular}
\end{table}

\begin{table}[H]
\centering
\caption{Extended cliff diagnostics on the three small datasets. Lower is better for Overall MAE and Pair Cliff MAE. Pair Sign is higher-is-better. Q4/Q1 and Cliff-NonCliff Gap are severity indicators rather than standalone quality scores; higher Q4/Q1 indicates greater cliff-induced error amplification. Positive \(\rho(\mathrm{error}, \mathrm{CliffScore})\) means that larger errors align with stronger cliff structure. Negative \(\rho(\mathrm{error}, T_{\max}^{\mathrm{train}})\) means that better training support tends to reduce error.}
\label{tab:small_extended_diagnostics}
\centering
\small
\setlength{\tabcolsep}{4pt}
\begin{tabular}{llrrrrrrrr}
\toprule
Dataset & Algorithm & Overall & Q4/Q1 & C-N Gap & Pair MAE & Pair Sign & $\rho_{e,s}$ & $\rho_{e,t}$ & $N_{\mathrm{pairs}}$ \\
\midrule
\multirow{5}{*}{ESOL}
& Uni-Mol   & 0.413 & 1.177 &  0.011 & 0.336 & 1.000 & -0.062 & -0.104 &  41 \\
& GotenNet  & 0.448 & 1.209 &  0.018 & 0.533 & 1.000 & -0.024 & -0.088 &  41 \\
& ViSNet    & 0.421 & 1.067 & -0.064 & 0.432 & 1.000 & -0.155 & -0.149 &  41 \\
& MoleculeFormer & 0.419 & 1.388 &  0.009 & \textbf{0.307} & 1.000 & -0.093 & -0.159 &  41 \\
& EMPP      & \textbf{0.407} & 1.085 & -0.037 & 0.467 & 1.000 & -0.048 & -0.062 &  41 \\
\midrule
\multirow{5}{*}{FreeSolv}
& Uni-Mol   & 0.485 & 1.248 & -0.121 & 0.363 & 1.000 &  0.146 &  0.045 &  19 \\
& GotenNet  & \textbf{0.358} & 0.878 & -0.341 & 0.664 & 1.000 &  0.051 & -0.078 &  19 \\
& ViSNet    & 0.392 & 0.891 & -0.311 & 0.523 & 1.000 & -0.034 & -0.346 &  19 \\
& MoleculeFormer & 0.451 & 1.079 & -0.257 & \textbf{0.264} & 1.000 &  0.003 & -0.283 &  19 \\
& EMPP      & 0.547 & 0.719 & -0.171 & 0.608 & 1.000 & -0.067 & -0.215 &  19 \\
\midrule
\multirow{5}{*}{Lipophilicity}
& Uni-Mol   & 0.506 & 1.783 &  0.108 & 0.985 & 0.969 &  0.156 & -0.104 & 127 \\
& GotenNet  & 0.518 & 1.367 &  0.066 & 0.949 & \textbf{0.976} &  0.144 & -0.009 & 127 \\
& ViSNet    & \textbf{0.475} & 1.521 &  0.080 & \textbf{0.876} & 0.929 &  0.154 & -0.038 & 127 \\
& MoleculeFormer & 0.541 & 1.771 &  0.105 & 0.912 & 0.969 &  \textbf{0.191} & -0.104 & 127 \\
& EMPP      & 0.617 & 1.446 &  0.071 & 1.007 & 0.898 &  0.109 & -0.105 & 127 \\
\bottomrule
\end{tabular}
\end{table}

% \begin{ack}
% This work was supported by Wuhan University Computer Laboratory.
% \end{ack}

%%%%%%%%%%%%%%%%%%%%%%%%%%%%%%%%%%%%%%%%%%%%%%%%%%%%%%%%%%%%

\clearpage
\bibliographystyle{plainnat}
\bibliography{refs}

\end{document}